\documentclass[11pt, a4paper, logo, copyright, nonumbering]{astribot}
\usepackage[authoryear, sort&compress, round]{natbib}
\usepackage{dblfloatfix}
\usepackage{ulem}
\usepackage{caption}
\usepackage{dramatist}
\usepackage{xspace}
\usepackage{pifont} % http://ctan.org/pkg/pifont
\usepackage{multirow}
\usepackage{tcolorbox}
\usepackage{xltabular}
\usepackage{longtable}
\usepackage{hyperref}
\usepackage{array}      % for vertical centering in cells
\usepackage{amsfonts}
\usepackage{amsmath}
\usepackage{amssymb}
\usepackage{lineno}
\usepackage{multirow}
\usepackage{adjustbox}
\usepackage{siunitx}
\usepackage{mathtools}
\usepackage{multirow}
\usepackage[bottom]{footmisc}
\usepackage{fontawesome5}
\usepackage{subcaption}
\usepackage{setspace}
\usepackage{lipsum} 
\usepackage{multicol}
\usepackage{annotate-equations}
\usepackage{pdfpages}
\usepackage{threeparttable}
\usepackage{makecell}
\usepackage{xcolor}
\usepackage{epigraph}
\usepackage{longtable}

\tcbset{textmarker/.style={
        halign=center,parbox=false,boxrule=0mm,boxsep=0mm,arc=0mm,
        outer arc=0mm,left=1mm,right=1mm,top=7pt,bottom=7pt,
        toptitle=1mm,bottomtitle=1mm,oversize,}}
\newtcolorbox{hintBox}{textmarker,
    colback=yellow!10!white}
\newtcolorbox{importantBox}{textmarker,
    colback=red!10!white}
\newtcolorbox{noteBox}{textmarker,
    colback=green!10!white}
\definecolor{shadeblue}{RGB}{74, 111, 255}
\definecolor{shadered}{RGB}{255, 107, 107}
\definecolor{shadegreen}{RGB}{76, 175, 80}

\makeatletter
\def\@BTrule[#1]{%
  \ifx\longtable\undefined
    \let\@BTswitch\@BTnormal
  \else\ifx\hline\LT@hline
    \nobreak
    \let\@BTswitch\@BLTrule
  \else
     \let\@BTswitch\@BTnormal
  \fi\fi
  \global\@thisrulewidth=#1\relax
  \ifnum\@thisruleclass=\tw@\vskip\@aboverulesep\else
  \ifnum\@lastruleclass=\z@\vskip\@aboverulesep\else
  \ifnum\@lastruleclass=\@ne\vskip\doublerulesep\fi\fi\fi
  \@BTswitch}
\makeatother

\addto\extrasenglish{
}

 {\begin{list}{}%
         {\setlength{\leftmargin}{#1}}%
         \item[]%
 }
 {\end{list}}
 
\renewcommand{\today}{}
\newcommand{\model}[0]{\mbox{Lumo-2}\xspace}

\title{Towards Predictive, Aligned, and Scalable Robot Learning
}

\author[*]{
Astribot Team
\\
\small
\texttt{research@astribot.com}
\\
\vspace{2em}
\small
Project Page: \href{https://www.astribot.com/research/Lumo2}{www.astribot.com/research/Lumo2}
\\
\vspace{1em}
\small
Author List in \hyperref[sec:contribution]{Contributions}
}

\begin{abstract}
Learning, at its core, extends beyond memorization to the ability to reason and to approach novel problems by navigating a space of possibilities. 
In this work, we introduce \model, a latent world-action model that generates actions by reasoning over world dynamics in latent space. The learned latent world dynamics capture physically grounded visual transitions, naturally encoding action-inducible future possibilities and providing a unified substrate for cross-modal alignment. This formulation enables \textbf{predictive reasoning} akin to world modelling, while remaining lightweight and focused on the evolution of physical dynamics relevant to control.
Central to our approach is the hypothesis that action generation quality is governed by the geometry of the latent space. We observe that standard reconstruction-based tokenization objectives for action induce representations biased toward low-level signal fidelity, prone to misalignment between reconstruction quality and downstream control performance. To address this limitation, we propose a multi-stage \textbf{modality pre-alignment} strategy, in which action representations are progressively aligned with latent world dynamics, vision, and language. This process enforces cross-modal consistency, promotes abstraction, and induces a semantically structured latent space conducive to predictive reasoning and enables \textbf{improved scaling properties}.
We provide a systematic empirical study of latent world modelling and modality alignment, analyzing their roles in scaling laws and out-of-distribution generalization. Our results demonstrate that \model achieves consistent gains over strong vision-language-action (VLA) and world-action model (WAM) baselines, with pronounced improvements in  challenging real-world tasks that require temporal reasoning, physical understanding, or high control complexity such as long-horizon and dexterous manipulation. These findings suggest that structured multi-modal alignment, coupled with predictive reasoning, is a fundamental principle for advancing generalizable embodied intelligence.
\end{abstract}

\begin{document}
\maketitle

\begin{figure}[t]
  \centering
  \includegraphics[width=\linewidth]{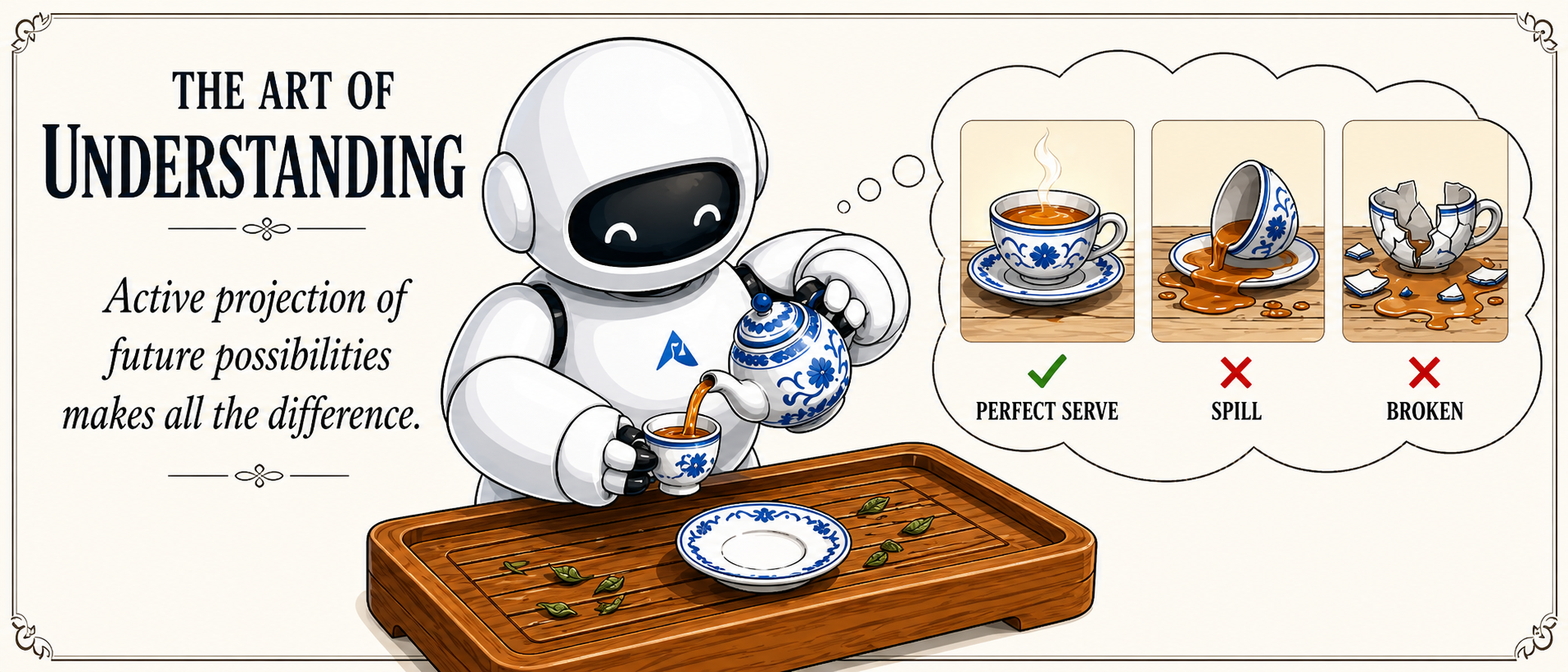}
   \label{fig:teaser}
\end{figure}

\section{Introduction}

\epigraph{``As projecting, understanding is the mode of being of Da-sein in which it \textit{is} its possibilities as possibilities."}{--- \textup{Martin Heidegger}, Being and Time
%, translated by Joan Stambaugh
}

As Martin Heidegger argues, human existence (Da-sein) is fundamentally defined by living in and through possibilities. Understanding is an active \textit{projection} of oneself into what one can be, where these possibilities remain inherently open.
In embodied settings, true robotic intelligence should similarly extend beyond memorizing experience (i.e., training data). It requires the capability to actively project toward future physical possibilities and to ground action in a coherent, aligned representation of perception, language, and physical dynamics.

In our prior work, Lumo-1~\citep{lumo1} employs explicit structured textual reasoning to guide action generation through coarse-to-fine planning. While effective, this formulation suffers from limited flexibility, poor scalability, and high inference latency. In this work, we introduce \model, which replaces explicit reasoning with latent reasoning. Specifically, we construct a latent world dynamics space to serve as an implicit reasoning bridge, reducing token complexity while providing a higher-capacity representation for capturing rich spatiotemporal dynamics and action-relevant dependencies. This latent space is trained to be physically grounded, naturally forming a structured space of action-inducible possibilities to which a learning agent, such as a robot, can align - akin to possessing an internal world model. 

Recent advances in vision–language–action (VLA) and world–action model (WAM) have shown promising progress towards endowing robots with general manipulation intelligence ~\citep{rt_2, pi_0, pi_0_5, pi0_7, gr00t_n1, dreamzero, bi2025motus, lingbotva}. However, while advances in the underlying vision–language and generative modeling have largely emphasized the quality of latent representations and modality alignment, the action modality in VLA and WAM models remains underexplored - particularly with respect to cross-modal alignment across modalities that exhibit fundamentally different information densities and statistical properties.
In this context, actions, as continuous control signals, must be generated in response to both visual observation - which provide environmental context but contain substantial irrelevant variation - and language, which encodes abstract semantics and task requirements. The introduction of latent world dynamics further exacerbates alignment challenges, as it lacks a natural correspondence with actions. Consequently, naive joint training over these heterogeneous modalities is prone to training instability, slow convergence, and poor generalization. Effective learning therefore requires explicit pre-alignment, which performs modality-aware structuring to bridge cross-modal gaps. To this end, we propose a multi-stage modality alignment paradigm, in which action representations are progressively aligned with latent world dynamics, visual observations, and language.

In this report, we introduce \model, a generalist latent world-action model for end-to-end robotic control. Conditioned on natural language instructions, visual observations, and robot proprioceptive states, \model projects future possibilities in latent space and generates viable actions to guide a whole-body bimanual robot toward the desired outcome. To fully leverage the understanding and generative capabilities of pre-trained foundation models, we treat the action modality as a first-class citizen, facilitating improved transfer and scaling behaviors. Concretely, we propose a \textbf{three-stage progressive training} paradigm, following a curriculum of increasing alignment difficulty, which enables the extraction of physically grounded latent world dynamics space while progressively optimizing action tokens and achieving multi-modal alignment:
\begin{enumerate}
    \item Stage~1 jointly learns to extract latent world dynamics and to optimize action tokenization via auto-encoding. Early \textbf{bimodal alignment} is achieved by leveraging latent world dynamics to provide global context for action reconstruction, which in turn grounds the learned world dynamics in the physical domain.
    \item  Stage~2 introduces semantic augmentation and multi-task learning to achieve \textbf{unified vision–language–action modality alignment}, while refining action tokenization to endow action tokens with semantic interpretability.
    \item Stage~3 builds upon the established cross-modal alignment and performs \textbf{large-scale co-training} across vision-language understanding, future projection, and action generation. For action generation, it explicitly promotes world modeling by anticipating future dynamics in latent space before generating actions.
\end{enumerate}

The proposed training paradigm is based on three core hypotheses: 
\begin{enumerate}
    \item Latent world dynamics, modeled from two observation timestamps, encode both action-relevant information and action-irrelevant redundancy. With appropriate constraints, this latent space can be distilled into compact, physically grounded future projection traces that could reliably guide action generation.
    \item Action sequences and latent world dynamics exhibit a dual relationship of containment and complementarity. Joint training can establish a causal linkage between world dynamics projection and action prediction, serving as a pre-alignment step.
    \item Action tokens can be elevated from purely kinematic representations to semantically enriched forms, enabling multi-modality pre-alignment for more scalable robot learning.
\end{enumerate}

We conduct in-depth analyses of latent world modeling and modality alignment, examining how latent space structuring influences the scaling behavior of generalist robot policies. We validate \model with extensive real-world experiments on three types of challenging tasks that require: (1) temporal reasoning, (2) physical understanding, and (3) high control complexity such as long-horizon and dexterous manipulation. Across all task categories, \model consistently outperforms state-of-the-art VLA and WAM baselines. Furthermore,  \model can be effectively fine-tuned using diverse, accessible data, such as egocentric human videos and VisionPro data.

\section{The \model Model}

\subsection{Embodied Understanding}
A central requirement for generalist robot intelligence is embodied reasoning - the capability to ground objects, relations, and knowledge in the physical world - which serves as a prerequisite for purposeful action generation. This capability is especially critical for robot manipulation, where generalizable and robust performance depends on understanding spatial structure, physical constraints, and the outcomes of interactions. Our goal is to develop a generalist model with embodied reasoning as its foundation, enabling strong generalization across diverse scenarios while maintaining rich multi-modal understanding. To endow the model with the broad visual and linguistic knowledge available in web-scale data, we build upon Qwen3.5-4B~\citep{qwen35blog} and further promote embodied, action-oriented reasoning.
To this end, we construct a large-scale dataset of high-quality vision–language data that extends beyond conventional corpora to better support embodied intelligence, and co-train it with robot data for action generation. The dataset is designed to jointly preserve general-purpose multi-modal understanding and strengthen robot-centric skills, with a focus on cognition, localization, and planning. 
Building upon the VLM data curation of Lumo-1~\citep{lumo1}, we further expand the dataset to emphasize learning physical dynamics from large-scale video data and richer spatial understanding. This enriched data composition is crucial for bridging perception and action, enabling the model to internalize abstract world understanding and develop reasoning capabilities oriented toward effective manipulation.

\subsection{Predictive Reasoning by World Modelling}
For embodied intelligence, reasoning should be predictive - informing future actions - while remaining flexible, expressive, and readily accessible. Therefore, we replace the explicit reasoning mechanism employed in prior work such as Lumo-1~\citep{lumo1} with latent reasoning. We instantiate this latent chain-of-thought paradigm through latent world dynamics that capture physically grounded visual transitions. The resulting latent space naturally encodes future possibilities inducible to action and serves as a unified substrate for cross-modal alignment.
This formulation enables predictive reasoning and bridges vision-language-action (VLA) and world-action model (WAM). While video-generative world models seek to predict future observations through dense pixel synthesis, our latent world-action model instead leverages world knowledge from large-scale VLM pre-training to infer a lightweight, future-relevant, action-centric latent representation.
Concretely, we adapt a pre-trained VLM to generate a latent world dynamics representation, denoted by $\phi$, which projects physically grounded world dynamics before low-level action generation. Given a robot demonstration dataset $\mathcal{D}$, our generalist policy $\pi_\theta$ is optimized via imitation learning, where each timestep $t$ consists of an observation $\mathbf{o}_t$ and a natural-language instruction $\ell$. Rather than directly predicting actions from observations, the model first infers latent world dynamics $\phi$, and then conditions action generation on this latent representation. The training objective maximizes the likelihood of the ground-truth action $\mathbf{a}_t$, or more generally an action chunk $\mathbf{a}_{t:t+H}$ over a horizon of $H$ timesteps:
\begin{equation}
    \max_\theta \, \mathbb{E}_{(\mathbf{a}_{t:t+H}, \mathbf{o}_t, \ell) \sim \mathcal{D}} 
    \; \log \big( \pi_\theta(\phi, \mathbf{a}_{t:t+H} \mid \mathbf{o}_t, \ell) \big).
\label{eq:optimization_objective}
\end{equation}

This formulation jointly optimizes latent world modeling and action generation, encouraging the model to ground its predicted actions in physically plausible future dynamics. Here, the observation $\mathbf{o}_t$ typically comprises multi-view visual observations together with the robot's proprioceptive state.

\subsection{Context Differentiation Towards Scalable Training}
\begin{figure}
  \centering
\vspace{-5em}
\includegraphics[width=\linewidth]{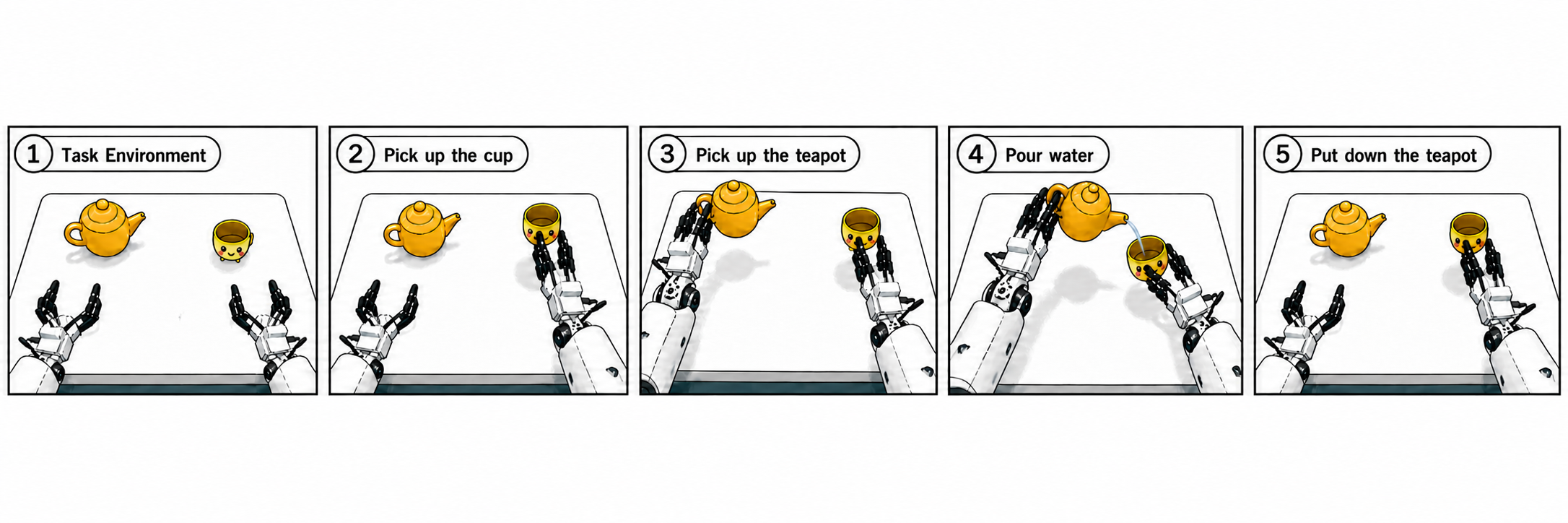}
\vspace{-4em}
   \caption{\textbf{Observation Ambiguity in Multi-Phase Task of Pouring Water.} (1) Panels 2 and 5 exhibit similar visual observations before and after pouring transparent water, and (2) panel 4 depicts an ongoing pouring action: in both scenarios a single-timestep observation is insufficient to determine the subsequent action.}
   \label{fig:context_differentiation}
\end{figure}

Modeling action generation from a single-timestep observation, as formulated in Eq.~\ref{eq:optimization_objective}, is inherently ill-posed. From a dynamical systems perspective, instantaneous observations are insufficient to uniquely determine the underlying system state, leading to partial observability and ambiguity in action inference. This myopia and ambiguity have two main consequences. First, such a formulation fails to capture temporally coherent control priors, analogous to the ``muscle memory'' exhibited in human manipulation. Second, real-world robotic interaction is inherently sequential and exhibits multi-modal temporal evolution: identical instantaneous observations may correspond to different underlying phases of execution. As illustrated in Fig.~\ref{fig:context_differentiation}, a typical ``pour water'' task involves multiple sources of ambiguity: (1) panels 2 and 5 share similar visual observations before and after pouring transparent water, and (2) the single-timestep observation in panel 4 does not indicate whether pouring should continue. Such ambiguities hinder policy learning by creating multi-modal observation-to-action relationships, in which identical observations can admit multiple candidate actions, yet only a small subset is task-consistent, whereas the remaining alternatives lead to erroneous or undesirable behaviors.

While one possible remedy is to introduce language annotations to disambiguate intermediate subtasks, this approach incurs significant annotation cost and limits scalability. Instead, we incorporate a short-term memory mechanism to approximate the latent system state. Specifically, we augment the observation with a temporal context buffer, denoted by $\boldsymbol{\omega}_{t-H':t}$, where $H'$ defines the context horizon. This memory provides a compact summary of recent history, enabling the model to better infer latent dynamics and resolve action ambiguity. Accordingly, the imitation training objective of Eq.~\ref{eq:optimization_objective} is extended to:
\begin{equation}
    \max_\theta \, \mathbb{E}_{(\mathbf{a}_{t:t+H}, \mathbf{o}_t, \ell, \boldsymbol{\omega}_{t-H^{\prime}:t}) \sim \mathcal{D}} 
    \; \log \big( \pi_\theta(\phi, \mathbf{a}_{t:t+H} \mid \mathbf{o}_t, \ell, \boldsymbol{\omega}_{t-H^{\prime}:t}) \big).
\label{eq:optimization_objective_w_history}
\end{equation}

\subsection{Progressive Action Tokenization and Modality Alignment}
Modality alignment is fundamental to multi-modal modeling. In vision–language models (VLMs), the projection module plays a pivotal role as the interface between perception and reasoning. Specifically, it transforms the visual features produced by the encoder into the embedding space of the large language model (LLM). By bridging these heterogeneous representations, the projection module aligns visual observations with the token space expected by the LLM, thereby enabling visual information to participate directly in the autoregressive reasoning and generation process. The same principle applies to generative models: as noted by previous works~\citep{Yao2025TowardsSP, Chen2025AligningVF}, for effective generation, the target latent space must go beyond a reconstruction-only objective and concisely capture high-level semantic information.

However, while adapting pre-trained VLMs and generative (world) models to form generalist robot policies, the action modality remains underexplored and lacks principled alignment with existing modalities. In practice, actions are typically modeled either as raw continuous signals, or as discretized tokens optimized for high-fidelity reconstruction, evolving from simple binning schemes~\citep{kim2024openvla} to more advanced approaches such as FAST~\citep{pertsch2025fast} and VAE-based autoencoders~\citep{liu2026rdt2,liu2025faster}. Reconstruction-driven objectives, however, prioritize low-level signal fidelity while ignoring that generation quality is governed by the geometry of the latent space; as a result, reconstruction-based action tokenization often leads to a misalignment between reconstruction accuracy and downstream control performance. To address this limitation, we propose a three-stage progressive training paradigm following a curriculum of increasing difficulty for action alignment: (1) pre-alignment with visual world dynamics, (2) semantic alignment with vision and language, and (3) joint world modeling and action generation with explicit causal structure. This design progressively elevates action representations toward higher-level semantics, enabling coherent alignment and a unified framework for cross-modal understanding and generation.

\begin{figure}[t]
  \centering
  \includegraphics[width=\linewidth]{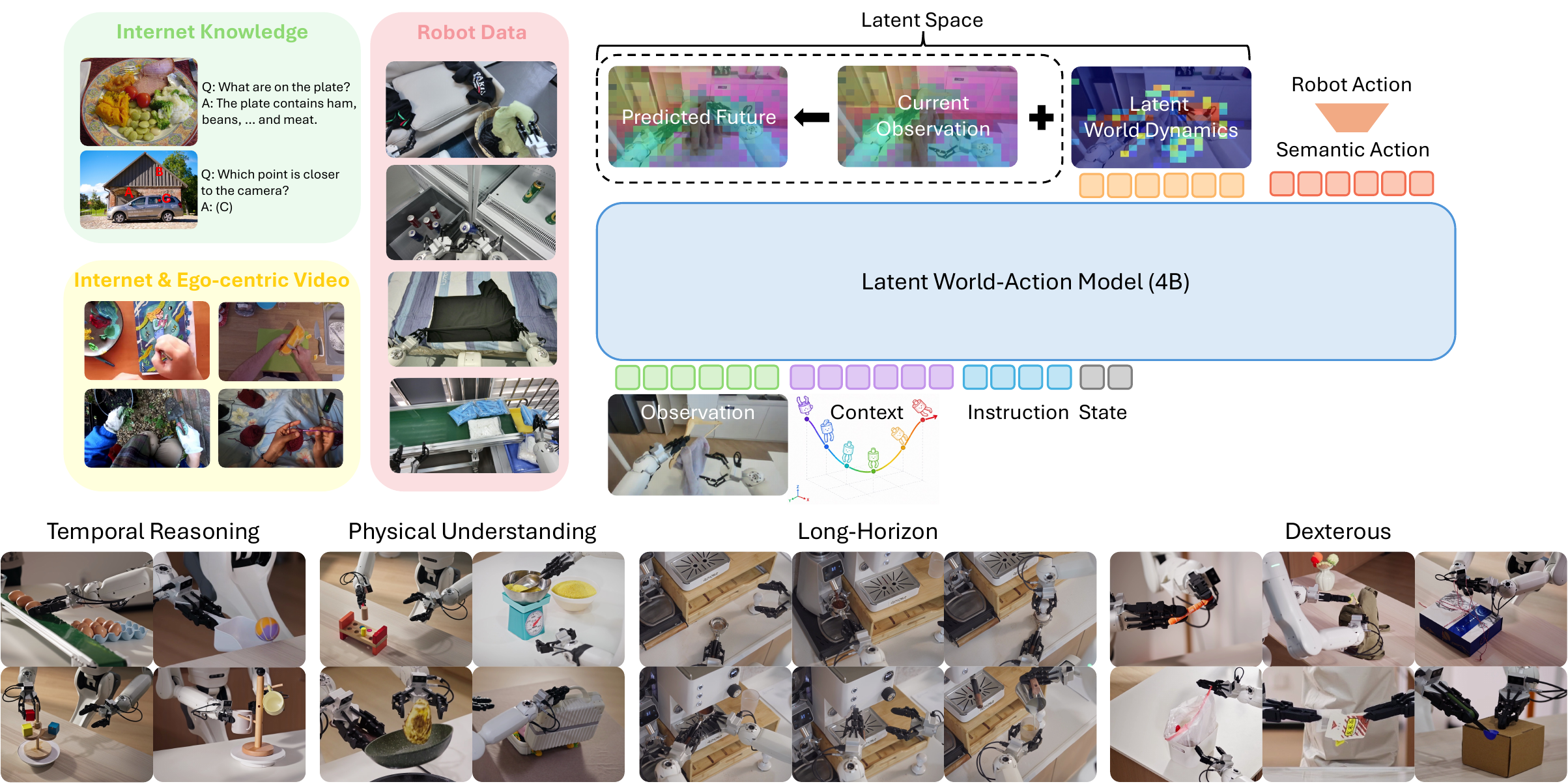}
     \caption{\textbf{\model: A Latent World-Action Model for Robotic Manipulation.}
    (1) \textbf{Co-training}: \model is co-trained on diverse vision-language data, in-the-wild videos, and robot action data, enabling robust generalization across scenes, objects, and world dynamics.
    (2) \textbf{Comprehensive Evaluation}: We conduct extensive real-world evaluations on challenging tasks that require temporal reasoning, physical understanding, and high control complexity, including long-horizon and dexterous manipulation.}
   \label{fig:model_architecture}
\end{figure}

\subsection{Model Architecture}
\model is an end-to-end, historical context–aware latent world-action model that jointly models the distributions over latent world dynamics, action chunks, and textual outputs for vision–language question answering. 
The action generation objective is defined in Eq.~\ref{eq:optimization_objective_w_history}, where latent world dynamics are modeled by $\pi_\theta(\phi \mid \mathbf{o}_t, \ell,\boldsymbol{\omega}_{t-H^{\prime}:t})$. Conditioned on this latent representation, action generation can be factorized into the following joint distribution:
\begin{equation}
\pi_\theta(\phi, \mathbf{a}_{t:t+H} \mid \mathbf{o}_t, \ell, \boldsymbol{\omega}_{t-H^{\prime}:t}) 
= \pi_\theta (\mathbf{a}_{t:t+H} \mid \mathbf{o}_t,\phi) \,
\pi_\theta(\phi \mid \mathbf{o}_t, \ell, \boldsymbol{\omega}_{t-H^{\prime}:t}),
\end{equation}

where the inference of low-level actions is conditioned solely on the latent world dynamics representation $\phi$. The high-level distribution over latent world dynamics is jointly trained with vision–language tasks to preserve and leverage the capabilities acquired during internet-scale VLM pretraining. Both the high-level distribution and the low-level action distribution are parameterized within a single, unified multi-modal transformer, as illustrated in Fig.~\ref{fig:model_architecture}.

\begin{figure}[t]
  \centering
  \includegraphics[width=\linewidth]{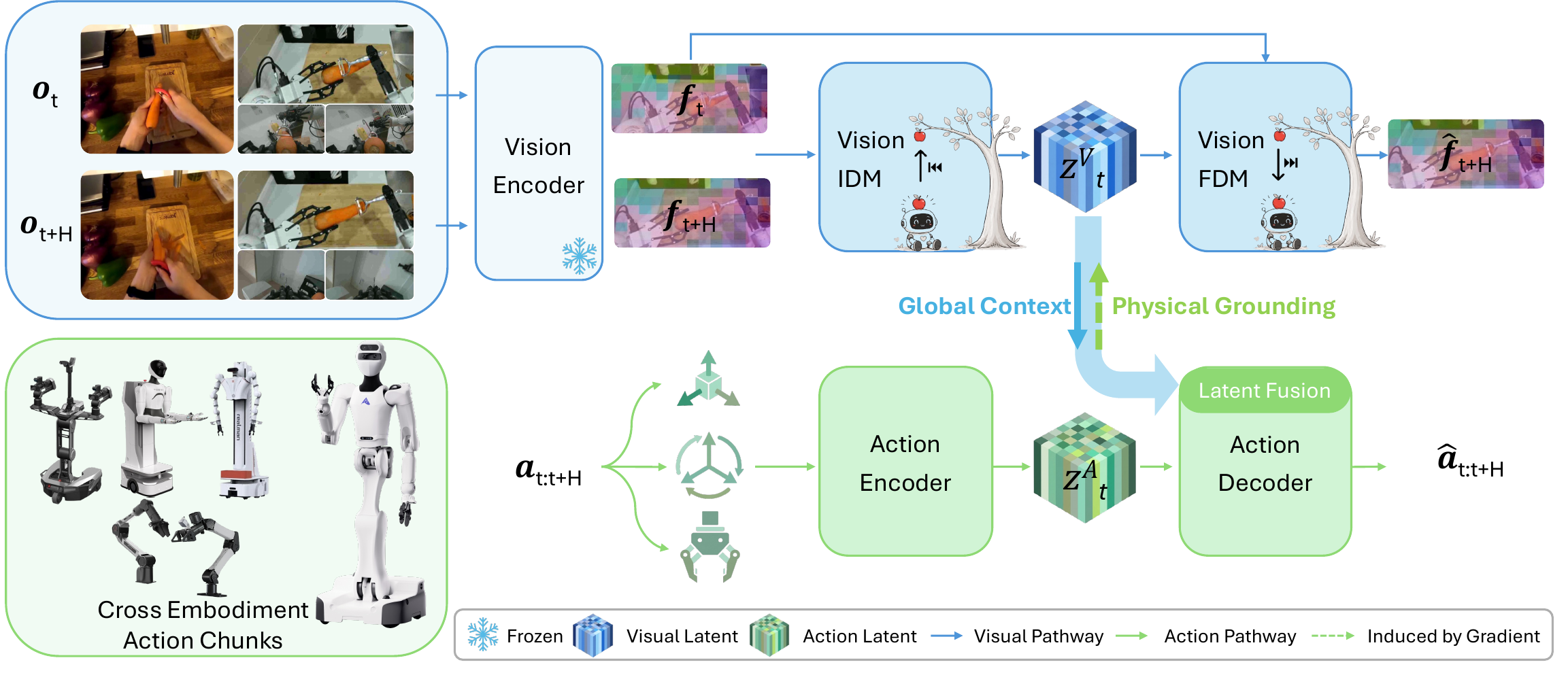}
   \caption{\textbf{Pre-align Action with Latent World Dynamics (Stage~1).} 
    We train a vision-based Inverse Dynamics Model (IDM) to encode world dynamics, denoted as $Z^V_t$, from inter-frame visual features extracted by a pre-trained image foundation encoder. By enforcing the visual representation to reconstruct corresponding actions, the model suppresses nuisance factors (e.g., illumination changes and background variations), producing a physically grounded latent space. At the same time, as $Z^A_t$ captures global action context, the action encoder is encouraged to model fine-grained action details coherently anchored to this global context.
    }
   \label{fig:stage1}
\end{figure}

\section{Training Recipe for \model}
We build \model by continuously training Qwen3.5-4B~\citep{qwen35blog} on a large-scale multi-modal corpus. First, we pre-align the action modality with latent world dynamics through joint training, enforcing the latent world dynamics to contribute to action reconstruction (Sec.~\ref{sec:stage1}). Second, we pre-align the action modality with vision and language via semantic augmentation and multi-task learning (Sec.~\ref{sec:stage2}). Finally, we train on latent reasoning–action data and embodied reasoning tasks to enable the model to systematically understand, project, and control (Sec.~\ref{sec:stage3}).

\subsection{Stage~1: Aligning Actions with Latent World Dynamics}
\label{sec:stage1}

The objective of Stage~1 training is to: (1) construct a physically grounded latent world dynamics space that captures action-inducible future possibilities, and (2) obtain an action encoder with global context awareness. We extract latent world dynamics from inter-frame visual observations, which provide a universal anchor for grounding kinematics across diverse embodiments, including human videos. 
We posit that inter-frame visual changes and corresponding actions exhibit a coupled relationship of inclusion and complementarity, implying that effective learning requires joint optimization for mutual enhancement. Specifically, visual changes encode temporal variations in the environment, reflecting dynamics induced by actions, but also capture action-irrelevant redundancy (e.g., lighting and background changes). While vision emphasizes global environmental context, action focuses on localized details and dynamics such as velocity and acceleration.
Training each modality in isolation therefore has inherent limitations: training vision alone via future frame reconstruction yields representations dominated by action-irrelevant visual factors, limiting its role as effective latent chain-of-thought for action guidance. Conversely, action alone auto-encoding yields context-disconnected representations, violating the premise that actions be conditioned on the environment.

To extract action-relevant world dynamics and to provide global context-awareness to action encoding, we adopt a co-training paradigm with feature fusion and bidirectional constraints to enable early alignment between world dynamics latent $\phi$ and action latent $A$. We enforce visual representations to contribute to action reconstruction, preserving visual components predictive of physical motion while suppressing nuisance signals. This yields a latent world dynamics space that is both physically grounded and aligned with the action modality, making it well-suited for subsequent multi-modal alignment. Concretely, we inject $\phi$ into the action decoder during action reconstruction:
\begin{equation}
\hat{\mathbf{a}} = D_A(A, \phi),
\end{equation}
which (1) refines action-relevant information in $\phi$, and (2) establishes a causal linkage between visual dynamics and action generation. This yields a synergistic mechanism:
\begin{equation}
\phi \rightarrow A \quad \text{(guidance)}, \qquad A \rightarrow \phi \quad \text{(regularization)},
\end{equation}
providing a strong foundation for subsequent multi-modal alignment and action generation with future envisioning.

\paragraph{Training Details.}

As illustrated in Fig.~\ref{fig:stage1}, we implement the proposed co-training framework with a dual-branch vector-quantized (VQ) architecture for joint modeling of latent world dynamics $\phi$ and action representations $A$, which enables early bi-modal alignment. The architecture consists of three core components: a vision auto-encoding branch, an action auto-encoding branch, and a cross-modal latent fusion module.

For the \textbf{vision auto-encoding branch}, we use frozen DINOv2~\citep{Oquab2023DINOv2LR} backbone for feature extraction. Inputs are 5-frame temporal sequences, each normalized to 224×224 and sampled every 8 frames at 30 FPS. For robot data, multi-view features are fused via cross-attention, with the head view used to query wrist views; for egocentric human videos, only the head view is used. Visual features are encoded by a causal spatial-temporal transformer~\citep{bu2025univla}, projected to a VQ codebook and optimized with a feature-level MSE loss for future feature reconstruction. For \textbf{action auto-encoding}, actions are first decomposed into semantic groups corresponding to the torso and bimanual arms, with each group further separated into translational, rotational, and gripper-motion components. Each component is independently projected to a unified channel dimension using dedicated linear layers, and the resulting features are temporally compressed from 32 to 4 frames using a Transformer AutoEncoder~\citep{parker2025scaling} to match the temporal resolution of the visual branch. The compressed action representations are then projected into a VQ codebook and optimized using an L1 reconstruction loss. For \textbf{latent fusion}, we use a multi-head attention strategy. The visual latent is expanded to match the 8 action groups and fused with the action latent to inject global environmental context into action decoding, realizing the bidirectional synergy between world dynamics guidance and action regularization.

We train our model on a heterogeneous, multi-source corpus comprising self-collected Astribot S1~\citep{gao2025towards} manipulation data, cross-embodiment robot datasets spanning diverse platforms - including bimanual humanoids such as AGIBot Genie-1~\citep{bu2025agibot}, Galaxea R1 Pro~\citep{jiang2025galaxea}, and RealSource RS-02~\citep{realsourceworld}, as well as tabletop dual-arm platforms like ARX, Agile X, and YAM. We further leverage egocentric human video datasets with rich world dynamics, including EgoTaskQA~\citep{Jia2022EgoTaskQAUH}, EPIC-KITCHENS~\citep{Damen2020TheED}, and Ego4D~\citep{Grauman2021Ego4DAT}. Astribot S1 data exhibit action speeds comparable to human demonstrations, while cross-embodiment datasets display significant variation in manipulation speed. As noted in~\citep{shi2026diversity}, such heterogeneity can hinder policy learning. To mitigate action-rate ambiguity, we apply a distribution debiasing method on cross-embodiment data. During training, we adopt a hybrid batching strategy: egocentric human data is processed solely through the vision branch, whereas robot data is jointly trained to optimize both vision and action branches.

The model is optimized with AdamW~\citep{loshchilov2017decoupled} using a base learning rate of $1 \times 10^{-4}$ and a weight decay of $1 \times 10^{-2}$, trained for 30,000 steps. For the VQ codebook parameters in both branches, we apply a $50\times$ learning-rate multiplier. To prevent codebook collapse and improve code utilization, all VQ codebooks use dead-code re-initialization. The learning rate follows a cosine decay schedule with a minimum value of $1 \times 10^{-6}$, combined with linear warmup over the first $5\%$ of training steps. We apply gradient clipping with a maximum norm of 1.0 and use bfloat16 mixed-precision training for numerical stability and computational efficiency. Training is conducted with a per-GPU batch size of 24 across 64 NVIDIA H100 GPUs. 

\begin{figure}[t]
  \centering
  \includegraphics[width=\linewidth]{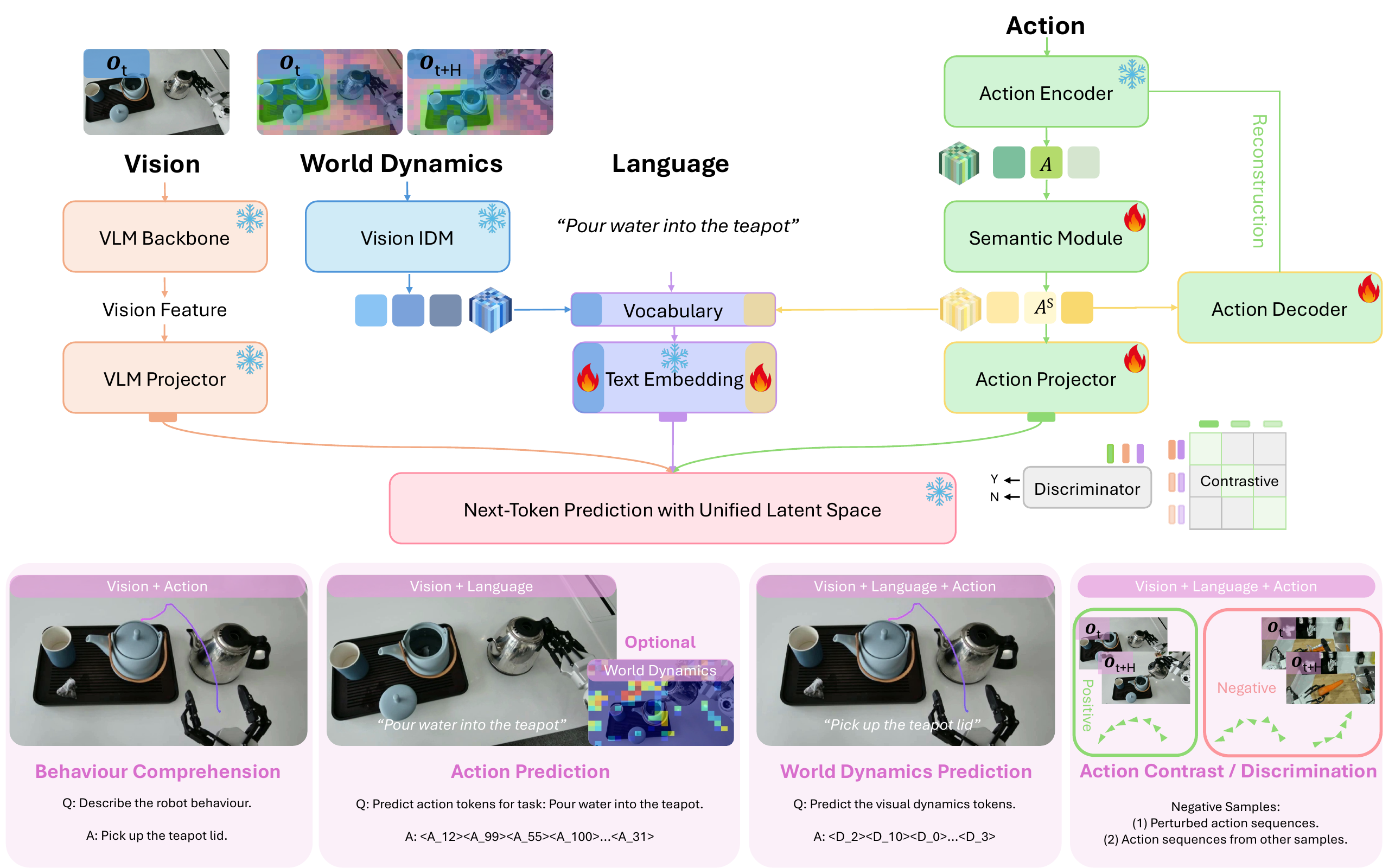}
   \caption{\textbf{Unified Modality Alignment (Stage~2).} 
We transform action tokens into semantically enriched representations and integrate vision and language modalities for unified alignment through multi-task, multi-modal learning.
    }
   \label{fig:stage2}
\end{figure}

\subsection{Stage~2: Aligning Actions with Vision and Language}
\label{sec:stage2}
In Stage~1, we learn physically grounded latent world dynamics and establish preliminary bimodal alignment between world dynamics and actions. However, the resulting action tokens primarily encode motion-specific patterns, lacking semantic content and effective associations with visual and language modalities. This leaves a generative modality gap that is commonly observed in current VLA and WAM models. Building on the dynamics–action alignment from Stage~1, Stage~2 explicitly incorporates visual and language modalities to transform the learned action tokens into semantically enriched representations. Specifically, we introduce a VLM backbone and a semantic extraction module, and employ multi-task learning together with an action projector to construct a unified semantic space for V–L–A alignment. This design endows action tokens with semantic interpretability while preserving fine-grained control details. It also reinforces the dynamics–action associations learned in Stage~1, thereby laying the foundation for end-to-end training in Stage~3.

\subsubsection{Architecture}
As illustrated in Fig.~\ref{fig:stage2}, Stage~2 focuses on semantic enhancement and modality alignment.

\paragraph{Semantic Module and Semantic Action Codebook.} We augment the Stage~1-trained action encoder with a semantic module, transforming the original action latent ($A$) into a semantically enriched latent ($A^{s}$) represented by a new codebook. This module leverages visual dynamics as a prior constraint, endowing action tokens with semantic interpretability and enabling alignment with the visual and language modalities. During training, the Stage~1 action encoder and its original codebook are frozen, while the semantic module, action projector, and action decoder are updated.

\paragraph{Feature Projector.} To accommodate new modalities and enable cross-modal alignment, we utilize separate projectors to map different modalities into a unified latent space. As the world dynamics latent is not updated, it is incorporated into the VLM vocabulary and adjusted via text embeddings. To ensure stable training and reduce alignment difficulty, only the projectors are trained, while the Stage~1 encoder and codebook, along with the VLM backbone and LLM, remain frozen.

\subsubsection{Multi-Task Learning}
To achieve semantic enrichment and cross-modal alignment, we employ multi-task learning with the following objectives:
\begin{itemize}
    \item \textbf{Action Reconstruction.} As the action reconstruction pathway from Stage~1 has been modified, we retain the action reconstruction task to ensure high-fidelity action reconstruction.
    \item \textbf{Behaviour Comprehension.} Generate natural language description given visual observation and action as input. This task constrains the joint representation of vision and action to accurately align to the language semantic space.
    \item \textbf{Vision–Language Guided Action Generation.} Predict discrete semantic action tokens given visual input and language instruction. This task ensures that vision and language effectively guide action generation.
    \item \textbf{Cross-Modal Prediction.} Generate actions given vision, language, and world dynamics, or generate world dynamics given vision, language, and actions. This task explicitly constrains the world dynamics and actions to be mutually informative.
    \item \textbf{Cross-Modal Action Contrast and Discrimination.} Construct contrastive and discriminative objectives across vision, language, and action modalities to enforce consistency in the learned representations. These constraints are applied at the embedding level: for contrastive loss, image, language, and action pass through their respective embedding mechanisms. For discriminative loss, an additional head is introduced to make predictions without passing through the VLM.
\end{itemize}

\paragraph{Training Details.}

We conduct Stage~2 cross-modal alignment training on our curated multi-robot demonstration dataset. We initialize the vision-language backbone with pre-trained Qwen3.5-4B~\citep{qwen35blog} weights, and load the Stage~1 pre-trained latent action model for visual and action encoding. To preserve the quality of pre-trained representations and stabilize training, we adopt a minimal fine-tuning strategy: only the newly introduced semantic module, feature projection layers, and additional loss-specific layers are trainable. We extend the vocabulary with 32 visual context tokens, 1024 semantic action tokens, and one action sequence placeholder token.
We use the AdamW~\citep{loshchilov2017decoupled} optimizer with a weight decay of 0.01. A layer-wise learning rate scheme with a cosine annealing schedule is applied: the base learning rate is $1 \times 10^{-4}$, with a $5\times$ scaling for the text embedding layer and $2\times$ scaling factor for layers associated with contrastive and discriminative objectives. We optimize a unified multi-task objective to achieve comprehensive cross-modal alignment, and train the model for a total of 12,000 steps across 64 NVIDIA H100 GPUs.

\subsection{Stage~3: Co-Training on VLM, Video, and Robot Data}
\label{sec:stage3}

\begin{figure}[t]
  \centering
  \includegraphics[width=0.7\linewidth]{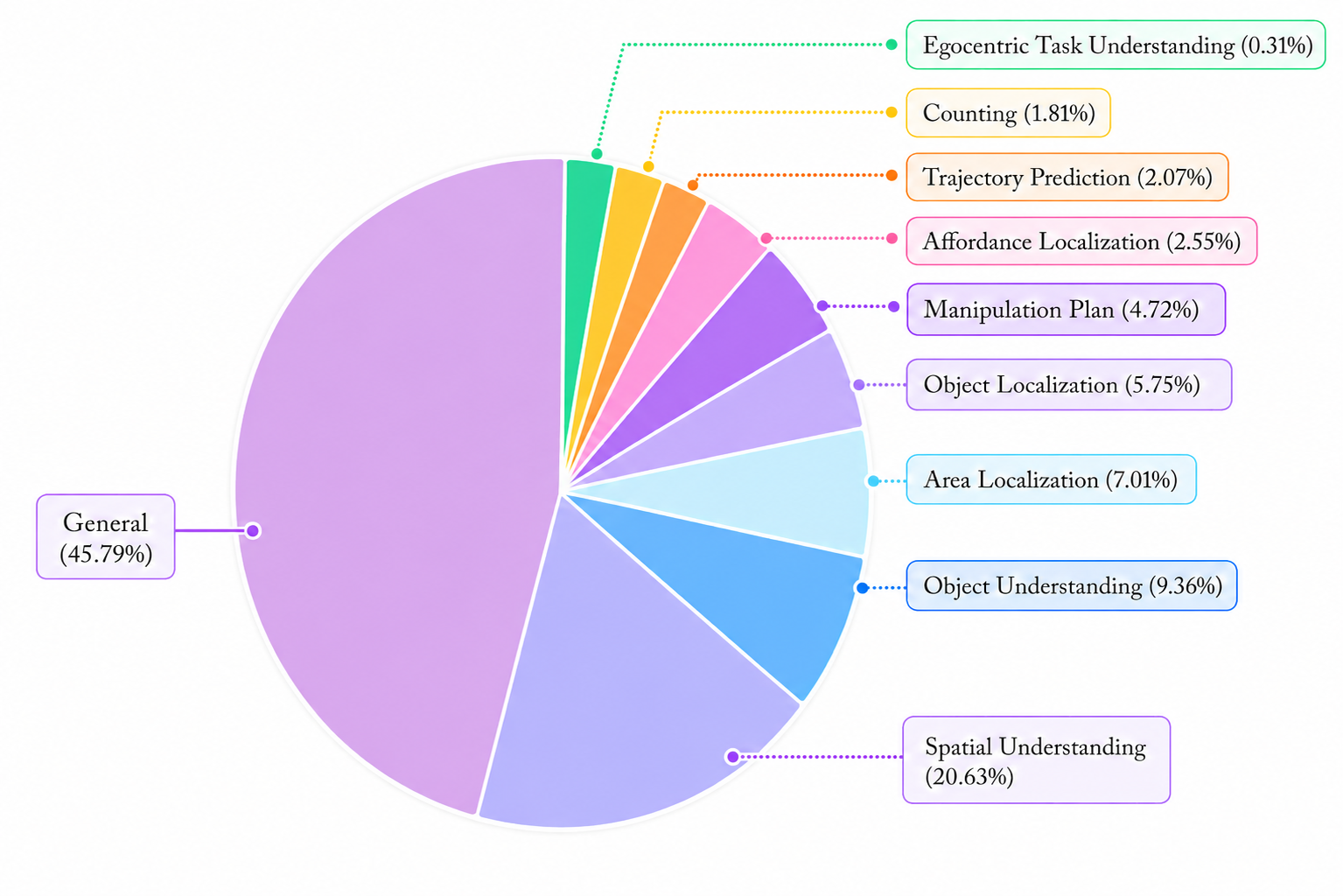}
   \caption{\textbf{Distribution of Data Mixture}. We curate a VLM dataset comprising roughly $53$ million samples that extend general multi-modal understanding with an emphasis on localization, cognition, and planning. }
   \label{fig:vlm_portion}
\end{figure}

\subsubsection{World Knowledge from VLM Data}
\begin{figure}[ht!]
  \centering
  \includegraphics[width=\linewidth]{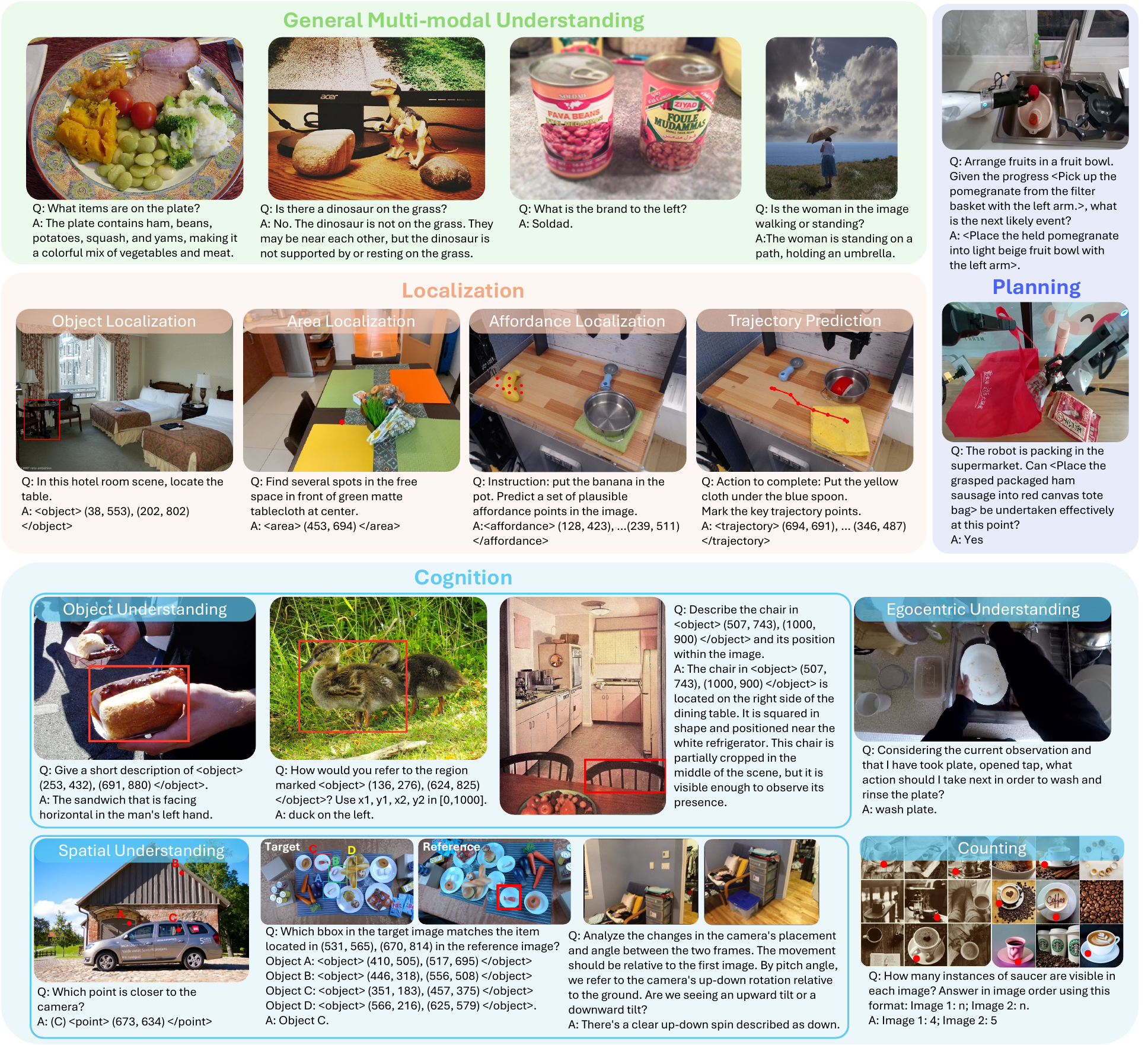}
   \caption{\textbf{Overview of Curated Vision-Language Data}. The curated dataset is designed to enhance core embodied reasoning abilities while preserving the general multi-modal understanding and reasoning capabilities of the pre-trained VLM. }
   \label{fig:vlm_data}
\end{figure}

Our goal is to develop a generalist robotic foundation model that preserves rich world knowledge and leverages it for future prediction. To this end, we construct a large-scale dataset consisting of high-quality vision-language data designed to: (1) enhance spatiotemporal understanding while retaining the broad capabilities of the pre-trained foundation model; (2) improve the modeling of physical dynamics and spatial relationships to support embodied reasoning; and (3) strengthen object recognition and localization capabilities, enabling accurate grounding of instructions to target objects and improving instruction-following performance. The dataset places particular emphasis on localization, cognition, and planning, while also incorporating diverse data for general-purpose multi-modal understanding. The distribution of the data mixture is illustrated in Fig.~\ref{fig:vlm_portion}. Representative examples from the dataset are shown in Fig.~\ref{fig:vlm_data}. Tab.~\ref{tab:vlm_datasets} lists all open-source raw datasets used in this work.

Compared to Lumo-1~\cite{lumo1}, \model incorporates several key improvements in data scaling and capability coverage:
\begin{enumerate}
    \item \textbf{Substantially expanded video data}, with a particular focus on datasets supporting general multi-modal understanding, spatial reasoning, egocentric understanding, and localization capabilities, inspired by Molmo2~\citep{Clark2026Molmo2OW}.
    \item  \textbf{A richer FineVision corpus}, utilized to better preserve and enhance the general-purpose vision-language capabilities inherited from the base foundation model.
    \item \textbf{Significantly expanded data for spatial understanding}, strengthening the model's ability to reason about object locations, spatial relationships, and embodied environments.
\end{enumerate}

\paragraph{General Multi-modal Understanding.} To preserve the general world knowledge of the underlying foundation model, we adopt FineVision~\citep{Wiedmann2025FineVisionOD} as the primary vision-language instruction tuning dataset. Compared with the Cambrian~\citep{tong2024cambrian} subset used in Lumo-1, which excluded several specialized domains such as OCR and scientific diagrams, FineVision provides substantially broader coverage and larger scale, comprising 24.3M samples. Prior work~\citep{Wiedmann2025FineVisionOD} has shown that FineVision yields stronger performance than Cambrian across a wide range of vision-language benchmarks. Building upon image-based instruction tuning, we further introduce a large-scale video understanding corpus to enhance the model's perception of spatiotemporal dynamics. Unlike static images, videos provide continuous temporal context, enabling the model to capture object motions, environmental changes, and task execution processes over time. The video data primarily focuses on general multi-modal understanding and dynamic scene comprehension, providing a stronger foundation for embodied reasoning.

\begin{table}[b!]
\centering
\setlength{\extrarowheight}{4pt}
\begin{tabularx}{\linewidth}{lX}
\toprule
Category           & Datasets \\
\midrule
General Understanding & EO-Data1.5M~\citep{Qu2025EO1AO}, FineVideo~\citep{Farre2024FineVideo}, FineVision~\citep{Wiedmann2025FineVisionOD}, LLaVA-Video~\citep{zhang2024videoinstructiontuningsynthetic}, ShareGPT4Video~\citep{chen2024sharegpt4video}, VCG+112k~\citep{Maaz2024VideoGPT+} \\
Localization  & ADE20K~\citep{zhou2019semantic}, COCOStuff~\citep{caesar2018coco}, EO-Data1.5M~\citep{Qu2025EO1AO}, PACO-LVIS~\citep{ramanathan2023paco}, PASCAL-Part~\citep{chen2014detect}, RoboRefit~\citep{lu2023vl}, Molmo2~\citep{Clark2026Molmo2OW}, RefSpatial~\citep{zhou2026roborefer}, RoboAfford~\citep{tang2025roboafford}, sharerobot~\citep{ji2025robobrain}, FSD~\citep{yuan2025seeing}, MolmoAct~\citep{lee2025molmoact} \\
Planning   & AgibotWorld~\citep{bu2025agibot}, EO-Data1.5M~\citep{Qu2025EO1AO}, Galaxea Open-World~\citep{jiang2025galaxea}, sharerobot~\citep{ji2025robobrain} \\
Cognition & DAM~\citep{lian2025describe}, Google Refexp~\citep{mao2016generation}, Osprey-724K~\citep{Osprey}, RefCOCO~\citep{su2025patchasdecodabletokenunifiedmultimodalvision}, VideoRefer-700k~\citep{yuan2025videorefer}, EO-Data1.5M~\citep{Qu2025EO1AO}, MessyTable~\citep{cai2020messytable}, MultiSPA~\citep{xu2025multi}, RefSpatial~\citep{zhou2026roborefer}, Sensenova-SI-800K~\citep{li2025controlvla}, VLM-3R~\citep{fan2026vlm}, VSI-590k~\citep{yang2025cambrian}, ViCA-322k~\citep{feng2025visuospatial}, Scannet~\citep{Dai2017ScanNetR3}, Scannet++~\citep{Yeshwanth2023ScanNetAH}, whatsup~\citep{kamath2023s}, Molmo2~\citep{Clark2026Molmo2OW}, EgoRe-5M~\citep{pei2025egothinkerunveilingegocentricreasoning}, QaEgo4d~\citep{barmann2022did}, Robo2vlm~\citep{chen2025robo2vlm}, EgoPlan-IT~\citep{chen2023egoplan} \\
\bottomrule
\end{tabularx}
\caption{\textbf{Overview of the multimodal data corpus organized by category.} We adopt publicly available datasets as the raw data source. Some datasets are assigned to multiple categories, and a portion of the data has been processed via filtering, cleaning, or additional construction.}
\label{tab:vlm_datasets}
\end{table}

\paragraph{Localization.} The localization data focuses on grounding language instructions to precise spatial locations and predicting structured coordinate outputs. Similar to Lumo-1~\citep{lumo1}, this category includes object localization, area localization, affordance localization, and trajectory prediction, while adopting a more standardized output format. Object localization requires the model to identify the bounding box corresponding to a referred object instance. Area localization focuses on detecting valid placement regions, such as free tabletop space. Affordance localization aims to identify task-relevant interaction regions, such as cup handles or buttons, given a manipulation instruction. Trajectory prediction requires the model to predict the corresponding two-dimensional end-effector coordinates or motion trajectory from visual observations and task instructions.

\paragraph{Planning.} The planning data focuses on robotic manipulation scenarios, where high-level instructions must be decomposed into executable action sequences. Most of the planning data originates from Lumo-1, and we further augment it with EO-Data1.5M~\citep{Qu2025EO1AO} dataset to increase task diversity and planning coverage. This data helps bridge the gap between abstract language instructions and concrete robotic behaviors, enabling more effective long-horizon manipulation planning.

\paragraph{Cognition.} The cognition data is constructed to improve the model's understanding of objects, spatial relationships, counting, and egocentric task execution. For object understanding, the model learns fine-grained semantic attributes of objects, including color, size, functionality, and orientation, by generating detailed descriptions given object bounding boxes. To strengthen spatial reasoning, we introduce data covering object size estimation, depth perception, relative spatial relationships, and viewpoint-aware reasoning. Inspired by SenseNova-SI~\citep{li2025controlvla}, we further construct datasets based on ScanNet~\citep{Dai2017ScanNetR3} and ScanNet++~\citep{Yeshwanth2023ScanNetAH} for distance estimation, relative position prediction, and perspective transformation understanding, enabling the model to reason not only from the current egocentric observation but also from alternative viewpoints. We additionally incorporate counting data derived from Molmo-2~\citep{Clark2026Molmo2OW} to improve numerical reasoning in visual scenes. Finally, we introduce large-scale egocentric task understanding data, including video sequences that capture task execution from a first-person perspective, helping the model understand how manipulation tasks evolve over time under different instructions.

\subsubsection{World Dynamics from Video Data}
\begin{figure}[ht!]
  \centering
  \includegraphics[width=\linewidth]{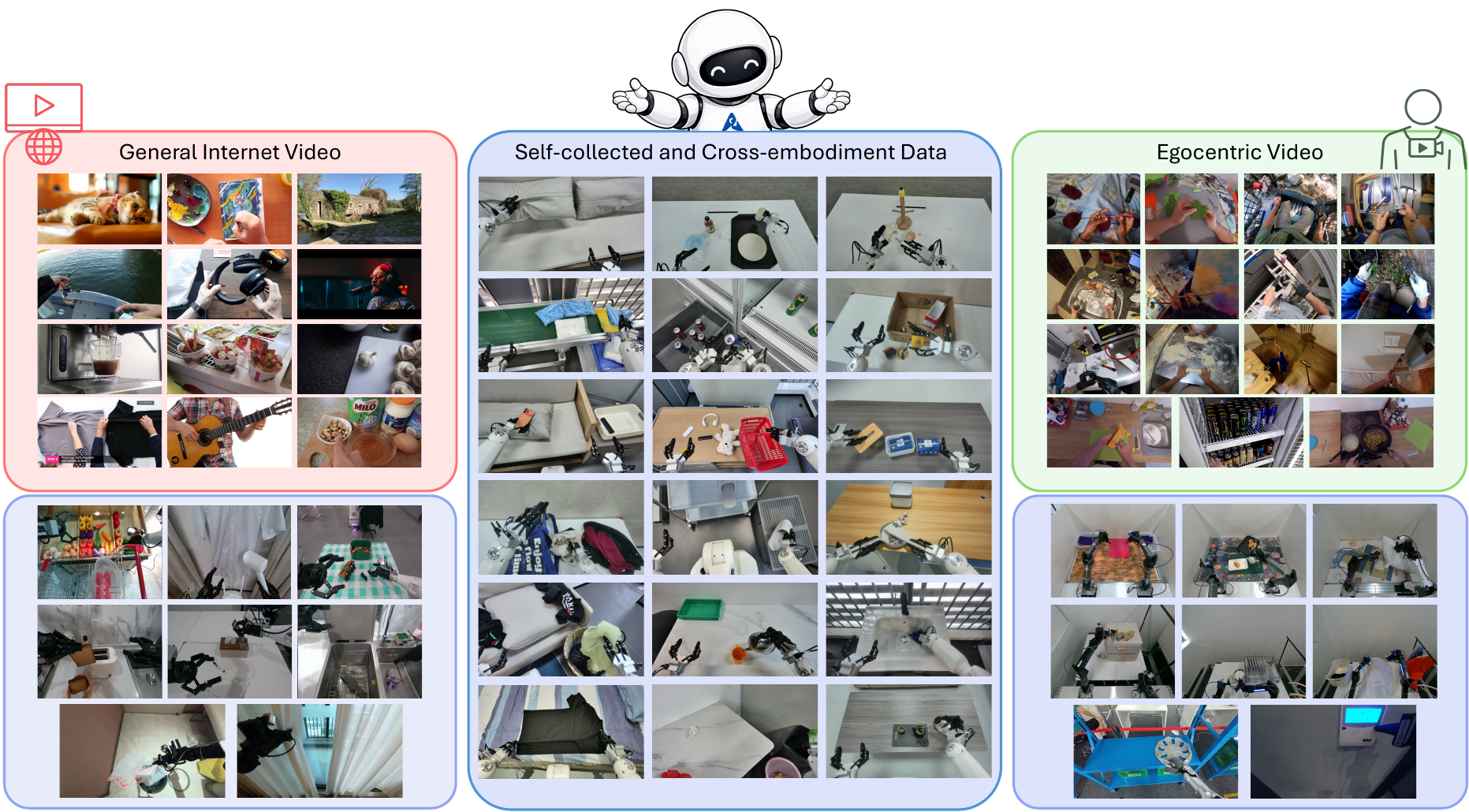}
   \caption{\textbf{Overview of Video Data Utilized to Learn Broad World Dynamics}. The curated dataset encompasses broad sources of general internet video, egocentric video, and self-collected and cross-embodied dataset. }
   \label{fig:video_data}
\end{figure}

Learning world dynamics can be efficiently achieved by observing the evolution of the natural world and the interactions between diverse embodiments and their environments. To this end, we construct a large-scale and diverse dataset that spans general internet videos, egocentric videos, and cross-embodiment robotic data, as illustrated in Fig.~\ref{fig:video_data}. By training on observations from these complementary perspectives, the model learns to predict world dynamics under a wide range of agents, environments, and interaction patterns.

\subsubsection{Context Differentiation} 
To achieve context differentiation towards scalable training, we propose a simple yet effective memory mechanism. Specifically, we maintain a fixed-length queue to store the robot's past actions. These historical actions are encoded via our semantic action encoder trained in Stage~2 and concatenated with other observation tokens to serve as the input to the generalist model as specified in Eq.~\ref{eq:optimization_objective_w_history}.
The short-term memory formed by historical actions enables the model to disambiguate perceptual aliasing. Furthermore, this memory paradigm offers two additional advantages: (1) Compared to image representations, action representations are significantly more compact, thereby improving both training and inference efficiency; (2) Historical actions and future actions share an identical encoder, ensuring semantic consistency in their encoded representations, which facilitates model learning.

\subsubsection{Block-wise Autoregression (BAR)} 
\begin{figure}[b!]
  \centering
  \begin{subfigure}{0.45\linewidth}
    \centering
    \includegraphics[width=\linewidth]{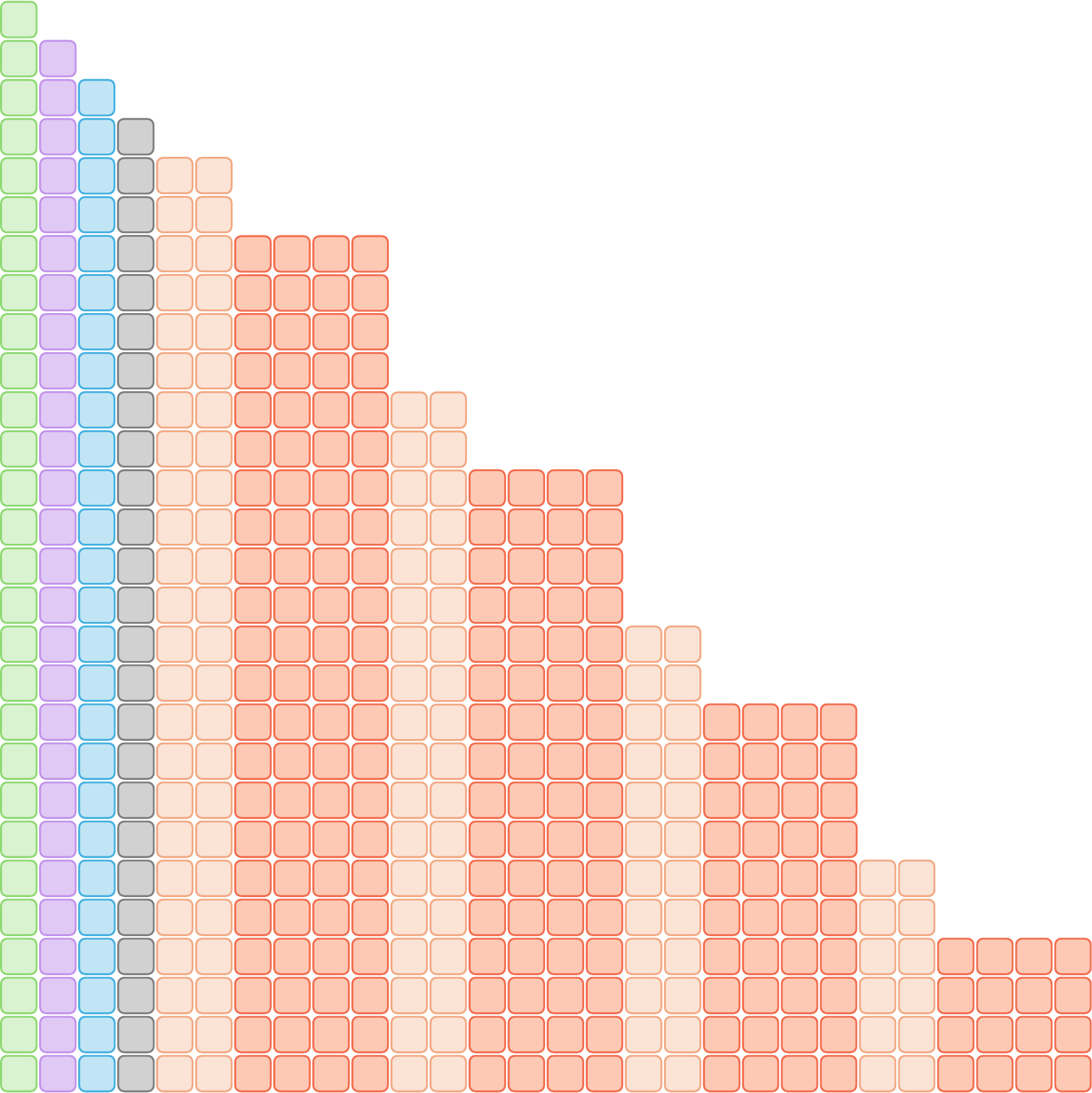}
    \caption{\textbf{BAR Causal Attention Mask.}}
    \label{subfig:bar_attention}
  \end{subfigure}
  \hfill
  \begin{subfigure}{0.45\linewidth}
    \centering
    \includegraphics[width=\linewidth]{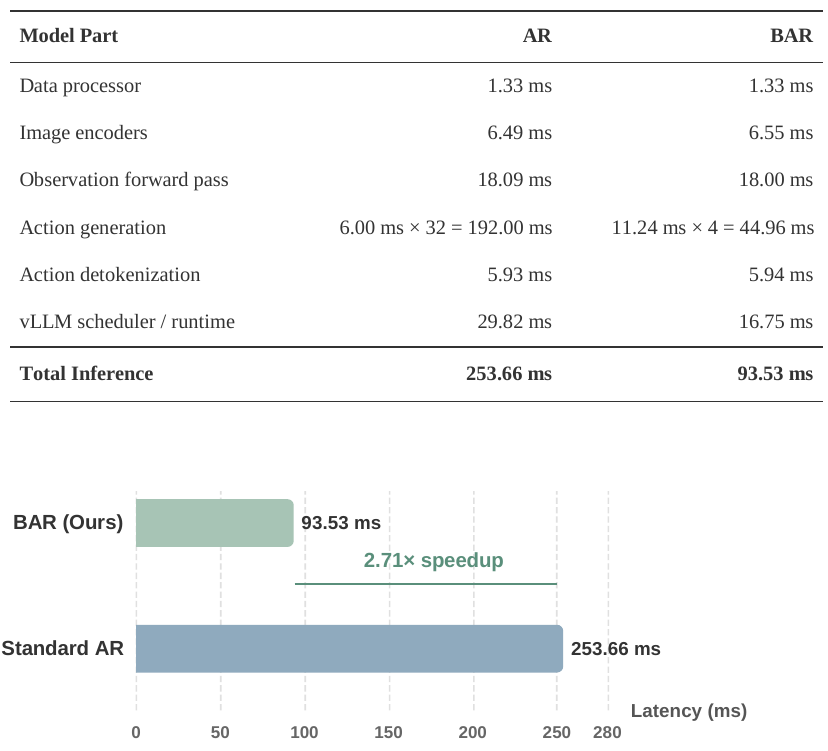}
    \caption{\textbf{Inference Latency Comparison.}}
    \label{subfig:bar_latency}
  \end{subfigure}
  \caption{\textbf{BAR Causal Attention Design and Inference Performance.} (a) Different colored regions correspond to vision, context, language, state, world dynamics, and action tokens, respectively. World dynamics and action tokens are arranged in alternating temporal order within each block, with block sizes configured as 4 and 8. (b) All latency measurements are conducted via vLLM on a single NVIDIA RTX 5090 GPU, and the proposed BAR decoding delivers a $2.71\times$ end-to-end speedup relative to standard AR decoding.}
  \label{fig:bar_overview}
\end{figure}

Standard policy training with autoregression uses a next-token prediction objective under teacher forcing for the subsequent action token. At inference, tokens are generated sequentially until the <eos> symbol and then decoded by the VQ decoder into a continuous trajectory $\mathbf{a}_{t:t+H}$. A major limitation of vanilla autoregression is its inference speed, which is critical for real-time robot control. It has been observed that many action codes are only weakly coupled across dimensions, as distinct action dimensions often correspond to independent physical semantics and exhibit heterogeneous distributions. Therefore, we adopt a block-wise objective that predicts the next block of tokens in a single forward pass. This approach is natural, as actions have already been partitioned into semantically meaningful blocks in Stage~1 training. We replace the standard causal mask with a block-wise causal mask that permits intra-block attention as illustrated in Fig.~\ref{subfig:bar_attention}. Block-wise autoregression (BAR) over discrete action codes enables efficient multi-token prediction while preserving autoregressive consistency, as similarly explored in Faster~\citep{liu2025faster}.

We further quantitatively compare the end-to-end inference latency between standard autoregressive (AR) decoding and the proposed BAR decoding strategy. All latency measurements are conducted using the vLLM~\citep{kwon2023efficient} inference framework on a single NVIDIA RTX 5090 GPU, with bfloat16 numerical precision and FlashAttention-2~\citep{dao2024flashattention} enabled for computational efficiency. As shown in Fig.~\ref{subfig:bar_latency} and the latency breakdown, BAR achieves an overall end-to-end inference time of 93.53 ms, corresponding to a $2.71\times$ speedup over standard AR decoding which takes 253.66 ms. The latency reduction primarily stems from two core modules. First, in the action generation stage, BAR compresses the original 32-step token-by-token autoregressive generation into 4-step block-wise parallel generation, reducing the latency of this stage from 192 ms to 44.96 ms and serving as the dominant source of speedup. Second, the vLLM scheduling and runtime overhead decreases from 29.82 ms to 16.75 ms, as the significantly reduced number of decoding steps lowers the extra costs of scheduling management and memory orchestration. The remaining modules, including data preprocessing, image encoding, and action detokenization, exhibit nearly identical latency across both schemes, verifying that the BAR mechanism introduces no additional computational overhead.

\paragraph{Training Details.}
In Stage~3 training, We build upon the Qwen3.5-4B~\citep{qwen35blog} backbone, initialized with Stage~2 vocab embeddings and additional special tokens. The model is trained for 120,000 steps on a mixed dataset using 160 NVIDIA H100 GPUs. The robot data subset covers VLA, VLW and VLWA task formats sampled with assigned weights, where V stands for vision, L stands for language, W stands for world dynamics, and A stands for action. For VLM data, we retain the standard AR training paradigm; for egocentric video data, we perform supervision under the VLW paradigm. To improve training efficiency, we adopt sequence packing with a maximum sequence length of 8{,}192 tokens and dynamic image resolution ranging from 50{,}176 to 307{,}200 pixels. Additionally, we apply a dropout rate of 0.5 to historical action inputs. We use the AdamW~\citep{loshchilov2017decoupled} optimizer with an initial learning rate of $1\times 10^{-5}$, a weight decay of 0.1 and gradient clipping at a norm of 1.0, paired with a cosine learning rate scheduler.

\section{Experiments}
We conduct extensive experiments to rigorously evaluate the multi-stage training paradigm and the performance of \model, focusing on the following key research questions:
\begin{itemize}
\item $\mathcal{Q}1$: Does \model effectively enhance embodied reasoning capabilities?
\item $\mathcal{Q}2$: Does latent world modeling provide an effective physical reasoning trace that improves out-of-distribution generalization?
\item $\mathcal{Q}3$: Can \model achieve effective few-shot adaptation on challenging tasks characterized by temporal reasoning complexity, physical reasoning demands, and high control and execution complexity?
\item $\mathcal{Q}4$: Can \model be trained with more accessible data? 
\item $\mathcal{Q}5$: Does modality pre-aligning stages effectively align the action modality with latent world dynamics, vision, and language in terms of reconstruction quality and semantic discriminability, and does such pre-aligning lead to improved scaling behavior?
\end{itemize}

\subsection{VLM Evaluation [$\mathcal{Q}1$]}

To verify the effectiveness of our curated VLM dataset, we conduct controlled training for approximately one epoch on the collected multimodal corpus. Except for the VLM backbone and training data, all other hyperparameter settings are nearly identical to those of Lumo-1 Stage~1. Furthermore, to examine whether the co-training stage effectively balances learning across different data modalities and mitigates catastrophic forgetting of pure vision-language capabilities, we compare the performance of \model Stage~3 (joint multi-modal co-training) against the \model VLM baseline (vision-language-only training).

Tab.~\ref{tab:vlm_benchmark} summarizes the evaluation results of \model and competing 4B-scale models on two categories of embodied benchmarks, namely embodied understanding and embodied location, with all tests performed via the lmms-eval~\citep{zhang2024lmmsevalrealitycheckevaluation} framework. For all evaluation runs, the visual input and decoding configurations are unified as follows: for both image and video inputs, the pixel count of visual features ranges from a minimum of $49 \times 32 \times 32$ to a maximum of $300 \times 32 \times 32$; videos are sampled at 2 FPS, and the total number of visual inputs per sample is limited to 32 (up to 32 images for multi-image tasks and up to 32 frames for video tasks); greedy decoding with a temperature of 0 is applied throughout generation.

\begin{table}[b!]
\centering
\setlength{\extrarowheight}{4pt}
\resizebox{\linewidth}{!}{%
\begin{tabular}{llccc|ccc}
\toprule
\textbf{Capability} & \textbf{Benchmark} & \textbf{\begin{tabular}[c]{@{}c@{}}Lumo-2\\ VLM\end{tabular}} & \textbf{\begin{tabular}[c]{@{}c@{}}Lumo-2\\ Stage 3\end{tabular}} & \textbf{\begin{tabular}[c]{@{}c@{}}Lumo-1\\ Stage 1\end{tabular}} & \textbf{\begin{tabular}[c]{@{}c@{}}Qwen-3.5\\ 4B\end{tabular}} & \textbf{\begin{tabular}[c]{@{}c@{}}RoboBrain-2.5\\ 4B\end{tabular}} & \textbf{\begin{tabular}[c]{@{}c@{}}Rynnbrain\\ 4B\end{tabular}} \\ \hline
 & BLINK-Depth & \cellcolor[HTML]{C2C8FC}\textbf{91.93} & 87.1 & 87.9 & 79.03 & 86.29 & \cellcolor[HTML]{DED9FA}88.7 \\
 & BLINK-Spatial & \cellcolor[HTML]{DED9FA}80.42 & \cellcolor[HTML]{C2C8FC}\textbf{86.01} & 77.6 & \cellcolor[HTML]{C2C8FC}\textbf{86.01} & 80.41 & 77.6 \\
 & BLINK-Mean & \cellcolor[HTML]{DED9FA}86.18 & \cellcolor[HTML]{C2C8FC}\textbf{86.55} & 82.4 & 82.52 & 83.35 & 83.15 \\
 & CV-Bench & 86.92 & \cellcolor[HTML]{DED9FA}87.45 & 86.36 & 84.84 & 86.54 & \cellcolor[HTML]{C2C8FC}\textbf{88.5} \\
 & Ego-Plan2 & 42.01 & \cellcolor[HTML]{DED9FA}47.46 & 36.49 & 37.77 & \cellcolor[HTML]{C2C8FC}\textbf{51.85} & 40 \\
 & VSIBench & \cellcolor[HTML]{C2C8FC}\textbf{52.34} & 48.56 & 27.61 & 21.73 & 32.84 & \cellcolor[HTML]{DED9FA}51.97 \\
 & MindCube & \cellcolor[HTML]{C2C8FC}\textbf{66.08} & \cellcolor[HTML]{DED9FA}54.93 & 25.54 & 39.95 & 28.77 & 53.95 \\
 & MMSI & \cellcolor[HTML]{DED9FA}36.4 & 35.2 & 26.9 & 30.8 & 29 & \cellcolor[HTML]{C2C8FC}\textbf{37.4} \\
\multirow{-9}{*}{\begin{tabular}[c]{@{}l@{}}Embodied\\ Understanding\end{tabular}} & ViewSpatial & \cellcolor[HTML]{DED9FA}53.08 & 51.15 & 37.75 & 45.47 & 41.14 & \cellcolor[HTML]{C2C8FC}\textbf{55.1} \\ \hline
 & Where2place & \cellcolor[HTML]{C2C8FC}\textbf{74} & 70 & 69.06 & 55 & \cellcolor[HTML]{DED9FA}73 & 65 \\
 & RefSpatialBench-Loc. & 56 & \cellcolor[HTML]{DED9FA}63 & 57 & 53 & \cellcolor[HTML]{C2C8FC}\textbf{65} & 59 \\
 & RefSpatialBench-Plc. & \cellcolor[HTML]{C2C8FC}\textbf{74} & 60 & 54 & 39 & \cellcolor[HTML]{DED9FA}62 & 53 \\
 & RefSpatialBench-Unseen & \cellcolor[HTML]{C2C8FC}\textbf{54.55} & \cellcolor[HTML]{DED9FA}49.35 & 38.9 & 40.26 & 42.86 & 36.4 \\
 & RefSpatialBench-Mean & \cellcolor[HTML]{C2C8FC}\textbf{61.52} & \cellcolor[HTML]{DED9FA}57.45 & 49.96 & 44.09 & 56.62 & 49.47 \\
 & ShareRobot-Aff. & 58.5 & 59 & - & 52.5 & \cellcolor[HTML]{C2C8FC}\textbf{67} & \cellcolor[HTML]{DED9FA}60.5 \\
\multirow{-7}{*}{\begin{tabular}[c]{@{}l@{}}Embodied\\ Location\end{tabular}} & ShareRobot-Traj. (DFD) $\downarrow$& \cellcolor[HTML]{DED9FA}0.2259 & 0.2386 & - & 0.428 & \cellcolor[HTML]{C2C8FC}\textbf{0.1587} & 0.342 \\
\bottomrule
\end{tabular}
}
\caption{\textbf{Performance of Lumo-2 on embodied related benchmarks.} The $~ \downarrow ~$ symbol indicates lower values are better, \colorbox[HTML]{C2C8FC}{dark purple} denotes the best result per row, and \colorbox[HTML]{DED9FA}{light purple} denotes the second-best.}
\label{tab:vlm_benchmark}
\end{table}

Compared with the foundational Qwen-3.5 4B~\citep{qwen35blog} model, \model VLM yields consistent and remarkable improvements across almost all tasks. For example, the MMSI~\citep{yang2025mmsi} score rises from 30.8 to 36.4, VSIBench~\citep{yang2025thinking} accuracy increases from 21.73 to 52.34, and ViewSpatial~\citep{li2025viewspatial} improves from 45.47 to 53.08. These pronounced performance gains empirically demonstrate the value of our curated large-scale embodied multimodal dataset, which effectively strengthens the model's spatial perception, long-horizon reasoning, and fine-grained localization abilities in real-world embodied scenarios. Against other state-of-the-art 4B-scale embodied models (RoboBrain-2.5 4B~\citep{tan2026robobrain} and Rynnbrain 4B~\citep{dang2026rynnbrain}), \model VLM delivers highly competitive results. It achieves the best performance on multiple benchmarks including VSIBench, MindCube~\citep{yin2025spatial}, and RefSpatialBench~\citep{duetal2024embspatial}; it also ranks first on the Where2place~\citep{yuan2024robopoint} task in the embodied location category, establishing its strong competence in embodied perception and spatial cognition.

We further investigate the effect of co-training by comparing \model Stage~3 with its VLM counterpart. The results reveal that after integrating action modality training, \model Stage~3 retains overall comparable performance on pure vision-language benchmarks, and even surpasses the VLM variant on tasks such as BLINK-Mean~\citep{Fu2024BLINKML}, CV-Bench~\citep{tong2024cambrian}, and Ego-Plan2~\citep{qiu2026egoplan}. This finding indicates that our multi-modal co-training scheme successfully balances the learning of vision-language knowledge and action signals, and does not induce severe catastrophic forgetting of visual-linguistic capabilities. Finally, \model Stage~3 consistently outperforms Lumo-1 Stage~1 (trained with pure vision-language objectives) across all reported metrics, and also shows clear advantages on the ShareRobot~\citep{ji2025robobrain} benchmark in both affordance understanding and trajectory prediction.

\subsection{Utility of Latent World Dynamics [$\mathcal{Q}2$]}
In this section, we investigate the utility of the learned latent world dynamics from three perspectives. First, we probe the structure and semantic properties of the learned latent dynamics space. Second, we evaluate its impact on generalization using a suite of generalizable pick-and-place tasks. Third, we assess its effectiveness on post-training manipulation tasks by comparing policies with and without latent world dynamics envisioning.

\subsubsection{Probing Latent World Dynamics}

We analyze the latent world dynamics learned in Stage~1 to evaluate whether they provide a suitable representation for predictive reasoning.

\begin{figure}[h]
  \centering
  \includegraphics[width=\linewidth]{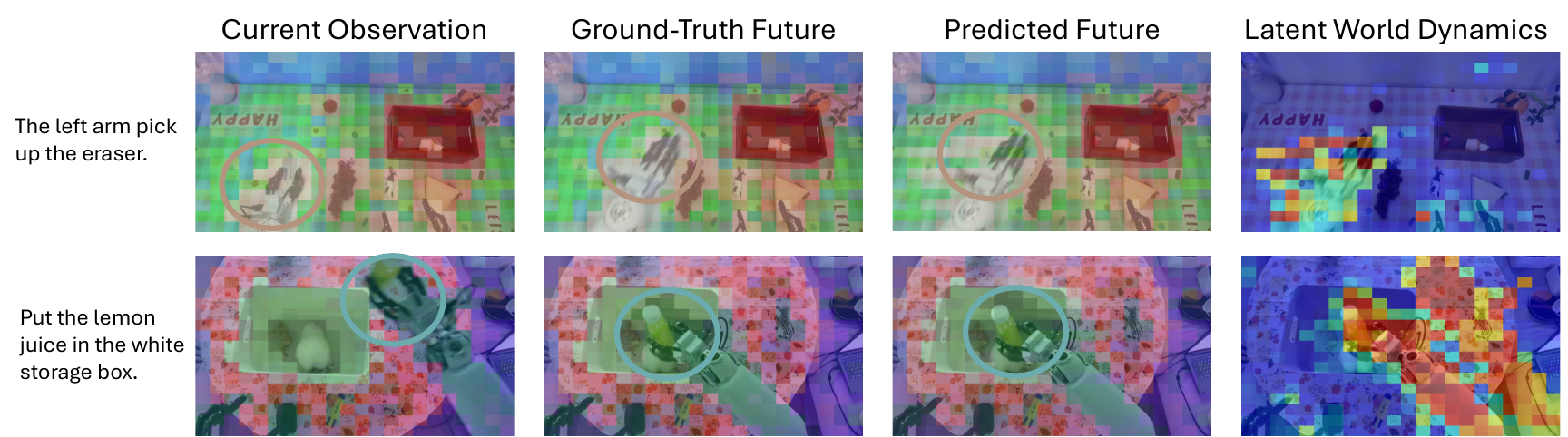}
   \caption{\textbf{Visualization of Latent World Dynamics and Observation Features.} The first three principal components of the feature representations are projected onto RGB channels for visualization. Latent world dynamics are visualized through the per-patch cosine distance between the predicted future and current observation features, highlighting regions where the model anticipates task-relevant future state changes.
    }
   \label{fig:latent_action_visualization}
\end{figure}

\paragraph{\textit{Does the latent world dynamics representation capture action-relevant future state evolution?}} To investigate whether the learned latent world dynamics encode task-relevant future changes, we visualize latent features using Principal Component Analysis (PCA), following the protocol of~\cite{Chen2026DIALDI}. Specifically, the first three principal components are mapped to RGB channels to provide an interpretable view of the latent feature space. As shown in Fig.~\ref{fig:latent_action_visualization}, the resulting color patterns reveal the spatial organization of the learned representation. For the task ``Put the lemon juice in the white storage box'' (Row 2), the predicted future representation exhibits strong visual similarity to the ground-truth future, particularly in regions associated with the manipulated object and target container (highlighted by circles). At the same time, noticeable differences emerge between the predicted future and the current observation in precisely those regions where object motion and interaction are expected to occur. The final column further visualizes these anticipated changes using the per-patch cosine distance between the predicted future and current observation features, where warmer colors indicate larger feature deviations. Across examples, the predicted future features consistently align with the ground-truth future while departing from the current state in manipulation-relevant regions. These results suggest that the latent world dynamics representation captures semantically meaningful future state evolution rather than simply preserving the present observation, providing action-relevant predictive context for downstream policy generation.

\begin{figure}[h]
  \centering
  \includegraphics[width=\linewidth]{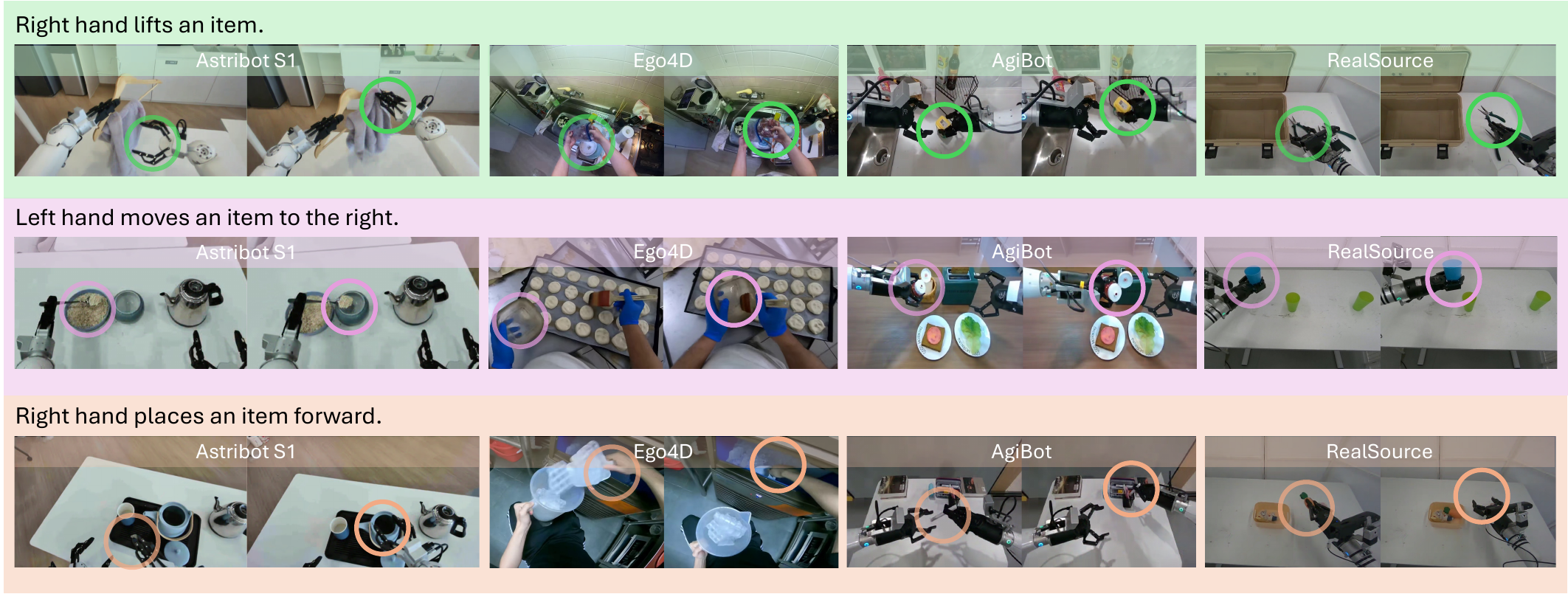}
   \caption{\textbf{Latent World Dynamics Align Action Semantics Across Embodiments.} Examples retrieved from clustered latent world dynamics tokens reveal emergent semantic alignment across diverse robot (Astribot S1, AgiBot, RealSource) and human (Ego4D) domains. Despite substantial differences in robot morphology, camera viewpoints, scene appearance, and end-effector designs, samples within each cluster exhibit consistent action semantics. Groups 1–3 correspond to lifting, transporting, and placing actions, respectively.
    }
   \label{fig:latent_action_correspondence}
\end{figure}

\paragraph{\textit{Does the latent world dynamics space encode semantic and embodiment-agnostic actions? }} To evaluate whether the learned latent world dynamics space exhibits cross-embodiment semantic consistency, we perform a cross-embodiment nearest-neighbor retrieval analysis. Following the Stage~1 training setup, we sample 32 frames from each video clip and extract one latent world dynamics every 8 frames, yielding 5 latent representations per clip. These latents are flattened and concatenated into a single clip-level representation, and cosine similarity is computed across clips from different embodiments. We then identify cross-embodiment pairs with cosine similarity greater than 0.99. Despite substantial differences in robot morphology, camera viewpoints, scene appearance, and end-effector designs, the learned latent action space consistently aligns semantically similar behaviors across embodiments. As shown in Fig.~\ref{fig:latent_action_correspondence}, clips depicting actions such as lifting, transporting, and placing objects are assigned highly similar latent action codes despite their markedly different visual appearances. This suggests that the learned latent world dynamics representation captures high-level, action-relevant semantics that transcend low-level visual appearance, mapping semantically similar behaviors performed by different embodiments to neighboring regions of the latent space. Such emergent cross-embodiment semantic alignment provides a strong foundation for subsequent latent action space alignment and action-semantic modeling.

\begin{figure}[h]
  \centering
  \includegraphics[width=\linewidth]{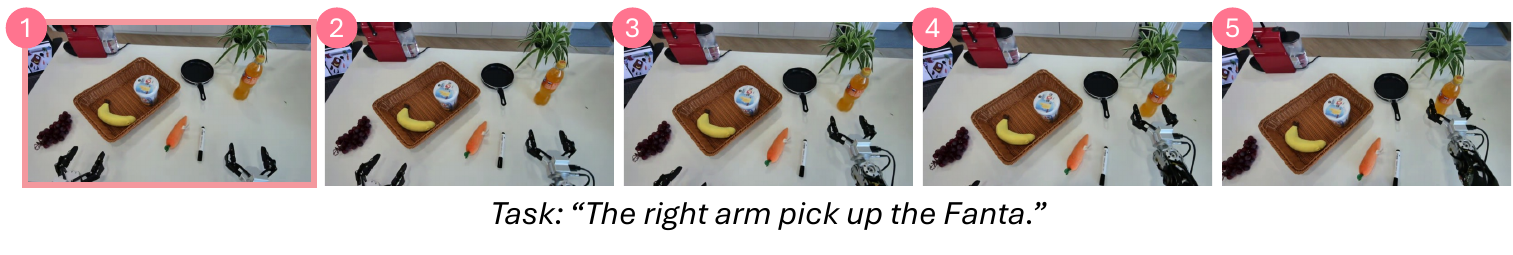}
   \caption{\textbf{Latent World Dynamics Encode Compact and Predictive Future Semantics.} Predicting task instructions from the initial observation and latent world dynamics achieves accuracy comparable to using five full observations, while substantially outperforming prediction from the initial observation alone.
    }
   \label{fig:task_prediction}
\end{figure}

\paragraph{\textit{Do latent world dynamics provide predictive guidance for action generation?}} 
We evaluate whether the learned latent world dynamics representation, $\phi$, encodes semantically meaningful and future-relevant information that can facilitate action generation, following the evaluation protocol proposed in~\citep{chen2025moto}. Specifically, we consider a task-inference benchmark in which the model predicts the underlying task instruction from different observation modalities, as illustrated in Fig.~\ref{fig:task_prediction}. The benchmark focuses on the early stage of manipulation, where multiple tasks share highly similar initial observations and the intended task is therefore deliberately ambiguous.
Using only DINO features extracted from the initial observation frame yields a classification accuracy of $43\%$. In contrast, incorporating DINO features from four subsequent frames improves accuracy to $94\%$, as the additional observations reveal task-disambiguating temporal cues. To assess the semantic content of $\phi$, we replace these four future observation frames with the corresponding latent world dynamics representations computed from consecutive frame pairs. Despite operating in a highly compressed latent space, this setting achieves $90\%$ accuracy, closely approaching the performance of the full five-frame DINO baseline. These results indicate that $\phi$ captures compact yet highly informative future-relevant semantics, effectively summarizing task-critical dynamics that are predictive of future behavior and useful for downstream action generation.

\subsubsection{Generalizable Pick and Place }
Following Lumo-1~\citep{lumo1}, we evaluate model instruction following and generalization capabilities on the task of generalizable pick and place with ``put A into/to B'' instruction, where A is the object and B is the target location. We evaluate 3 models, comparing \model with state-of-the-art VLA and WAM baselines: 
\begin{itemize}
\item $\boldsymbol{\pi_{0.5}}$: Fine-tuned on pick and place data collected on Astribot S1 for 10 epochs.
\item \textbf{Fast-WAM}: Fine-tuned on pick and place data collected on Astribot S1 for 3 epochs.
\item \textbf{\model}: \model directly evaluated after stage-3 training without further fine-tuning. 
\end{itemize}

We evaluate model capabilities under three settings: (1) \textbf{Basic}, (2) \textbf{Unseen Instructions}, and (3) \textbf{Unseen Objects}. In Basic, we use 60 training-seen objects to access the model's basic instruction-following ability. In Unseen Instructions, the model is prompted with instructions that demand higher-level conceptual understanding, such as spatial or semantic reasoning (e.g. ``put the [left coke] / [high-calorie drink] into the round woven basket'' ). In Unseen Objects, evaluation is performed on 105 novel objects that are absent from the training dataset, testing the model's ability to generalize to unseen items.

We evaluate model performance using two metrics: instruction-following rate (IFR) and task success rate (SR). The IFR reflects how accurately the robot identifies and approaches the object/location indicated in the instruction, while the SR measures whether the robot successfully completes the given instruction. Both metrics are reported as percentages, with higher values indicating better instruction comprehension and task execution capabilities. 
\begin{figure}[ht!]
  \centering
  \includegraphics[width=\linewidth]{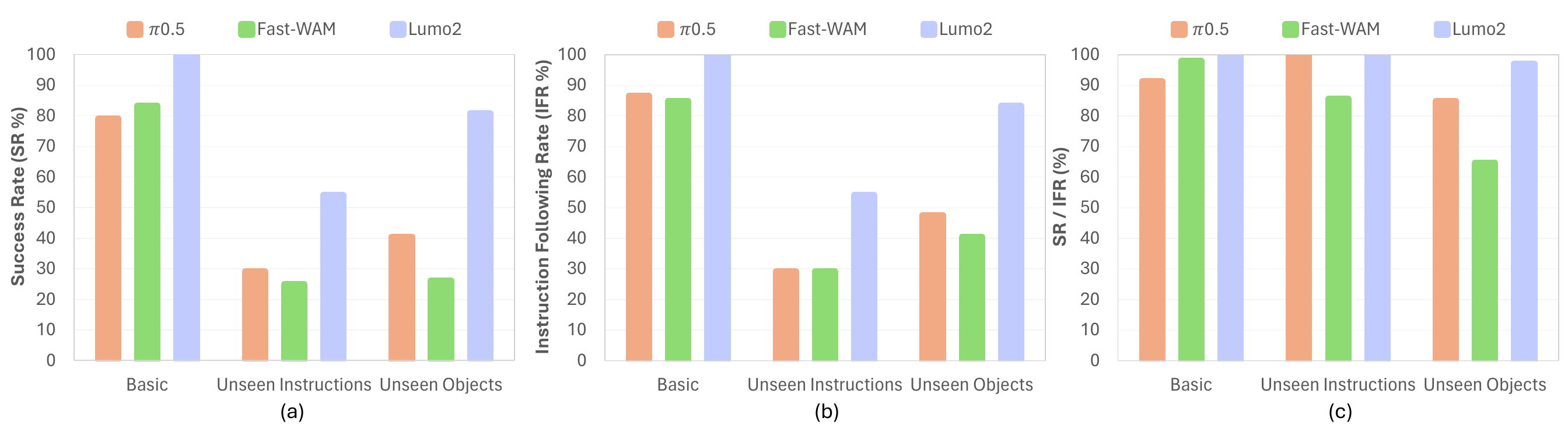}
   \caption{\textbf{Experiment Results of Generalizable Pick and Place}. As shown in (a) and (b), \model consistently outperforms the baseline $\pi_{0.5}$ and Fast-WAM across all three evaluation categories, with significant improvements on unseen instructions and unseen objects, which requires strong reasoning and generalization capabilities. \model exhibits enhanced action execution accuracy over $\pi_{0.5}$ and Fast-WAM as reflected by the IFR/SR metric in (c).}
   \label{fig:generalizable_pnp_results}
\end{figure}

As illustrated in Fig.~\ref{fig:generalizable_pnp_results}, \model consistently outperforms both $\pi_{0.5}$~\citep{pi_0_5} and Fast-WAM~\citep{yuan2026fast} across all three evaluation settings. The gains are particularly pronounced on unseen instructions and unseen objects, demonstrating strong reasoning and generalization capabilities. Moreover, \model achieves higher action execution accuracy than $\pi_{0.5}$, as evidenced by the SR/IFR metric. Under the unseen-instruction and unseen-object evaluations, both \model and $\pi_{0.5}$ substantially outperform Fast-WAM in terms of SR/IFR. We attribute this performance gap to the benefits of large-scale pre-training, which is absent in Fast-WAM and appears critical for acquiring robust and transferable manipulation skills.

\subsubsection{Training with Active Latent World Dynamics Projection}

We hypothesize that the latent world dynamics objective encodes task-relevant future state evolution and semantic information within the aligned latent space, which delivers disproportionate benefits for tasks that demand proactive reasoning about scene changes. In particular, temporal reasoning tasks—including reactive dynamic scene handling and predictive motion reasoning—rely critically on anticipating spatiotemporal evolution to enable timely, well-coordinated control, making them the primary categories expected to gain from explicit world dynamics modeling.

To validate this hypothesis, we conduct ablation studies based on Stage~3 models, comparing two fine-tuning paradigms: standard VLA fine-tuning with action supervision only, and full VLWA fine-tuning with active latent world dynamics projection enabled alongside action learning. We evaluate performance across the task taxonomy defined in Tab.~\ref{tab:task_categories} to quantify how world dynamics modeling contributes to different capability dimensions. We find that incorporating latent world dynamics yields consistent performance improvements, with the most pronounced gains concentrated in temporal reasoning categories. Detailed quantitative results are provided in Tab.~\ref{tab:vlwa_vs_vla}.

\begin{table}[htbp]
  \centering
  \begin{tabular}{ccc}
    \toprule
    Task & VLWA fine-tuning & VLA fine-tuning \\
    \midrule
    Collect Eggs from the Conveyor Belt & \textbf{100.00} & 92.00 \\
    Place Cubes on the Rotating Rack & \textbf{81.67} & 74.17 \\
    \bottomrule
  \end{tabular}
  \caption{\textbf{Success rate (\%) of representative temporal reasoning tasks under different training paradigms.}}
  \label{tab:vlwa_vs_vla}
\end{table}

\subsection{Task Post-Training [$\mathcal{Q}3$]}
To assess \model's effectiveness as a generalist robotic foundation model, we evaluate it on a diverse suite of challenging manipulation tasks spanning three broad capability dimensions: (1) temporal reasoning, (2) physical understanding, and (3) control complexity. These dimensions are further divided into several sub-categories, as summarized in Tab.~\ref{tab:task_categories}. Although many tasks require a combination of multiple capabilities, each task is assigned to the category that best reflects its primary challenge. All experiments are conducted on the high-performance bimanual mobile manipulation platform Astribot S1~\citep{gao2025towards}.

\begin{table}[h]
\renewcommand{\arraystretch}{1.25} 
\resizebox{\linewidth}{!}{%
\begin{tabular}{|c|c|c|}
\hline
\textbf{Category}                                                             & \textbf{Sub-Category}                                                    & \textbf{Task Properties and Policy Demands}                                                                                                                                                                                                                                          \\ \hline
\multirow{3}{*}{\begin{tabular}[c]{@{}c@{}}Temporal\\ Reasoning\end{tabular}} & \begin{tabular}[c]{@{}c@{}}Dynamic Scene\\  (Reactive)\end{tabular}      & \begin{tabular}[c]{@{}c@{}}Require time-critical responses to exogenously evolving states, where the feasibility of interaction is transient. \\ Act before objects move out of reach or the scene changes that invalidate the current plan, emphasizing rapid control.\end{tabular} \\ \cline{2-3} 
                                                                              & \begin{tabular}[c]{@{}c@{}}Motion Reasoning\\  (Predictive)\end{tabular} & \begin{tabular}[c]{@{}c@{}}Require anticipating future states through trajectory, velocity, and contact timing estimation.\\ Model spatiotemporal evolution to enable precise coordination, emphasizing prediction and timing-aware planning.\end{tabular}                           \\ \cline{2-3} 
                                                                              & Memory                                                                   & \begin{tabular}[c]{@{}c@{}}Involve sequential operations in which the same visual observation may correspond to different stages of execution.\\ Infer information from previous timesteps in order to infer the current phase and select appropriate actions.\end{tabular}          \\ \hline
\begin{tabular}[c]{@{}c@{}}Physical\\ Understanding\end{tabular}              & \begin{tabular}[c]{@{}c@{}}Physical \\ Understanding\end{tabular}        & \begin{tabular}[c]{@{}c@{}}Require predicting the consequences of interactions governed by physical laws, such as gravity and fluid dynamics. \\ Infer causal, often non-linear dynamics beyond purely kinematic reasoning.\end{tabular}                                             \\ \hline
\multirow{2}{*}{\begin{tabular}[c]{@{}c@{}}Control\\ Complexity\end{tabular}} & \begin{tabular}[c]{@{}c@{}}Long-Horizon \\ Execution\end{tabular}        & \begin{tabular}[c]{@{}c@{}}Involve extended sequences of interdependent actions with delayed rewards and compounding errors. \\ Maintain coherence over time, handle subgoals, and recover from intermediate failures, emphasizing robustness.\end{tabular}                          \\ \cline{2-3} 
                                                                              & \begin{tabular}[c]{@{}c@{}}Dexterous\\ Manipulation\end{tabular}         & \begin{tabular}[c]{@{}c@{}}Require high-precision control, often involving coordination, tight tolerances, and complex object geometries. \\ Handle high-dimensional control, contact stability, and sensitivity to small perturbations.\end{tabular}                                \\ \hline
\end{tabular}
}
\caption{\textbf{Breakdown of Evaluation Task Categories.} Evaluation tasks are broadly organized into three capability dimensions: (1) temporal reasoning, (2) physical understanding, and (3) control complexity. Each category is further divided into finer-grained sub-categories. The final column summarizes the key task characteristics and policy requirements.}
\label{tab:task_categories}
\end{table}

We adopt task-adaptive fine-tuning paradigms: static geometric manipulation tasks follow the standard VLA paradigm with action supervision only, while temporal reasoning and physical state prediction tasks employ the full VLWA paradigm with active latent world dynamics projection. For long-horizon tasks requiring sequential state tracking, we use two complementary temporal context schemes to mitigate perceptual aliasing: one encodes historical actions into compact tokens via our unified action tokenizer (detailed in the Stage 3 context differentiation mechanism); the other injects visual history traces by projecting bimanual end-effector trajectories onto the main camera view with time-encoded color gradients, achieving minimally invasive bidirectional temporal alignment across vision and language modalities.

\subsubsection{Task Suites}
\begin{figure}[p]
  \centering
  \includegraphics[width=0.95\linewidth]{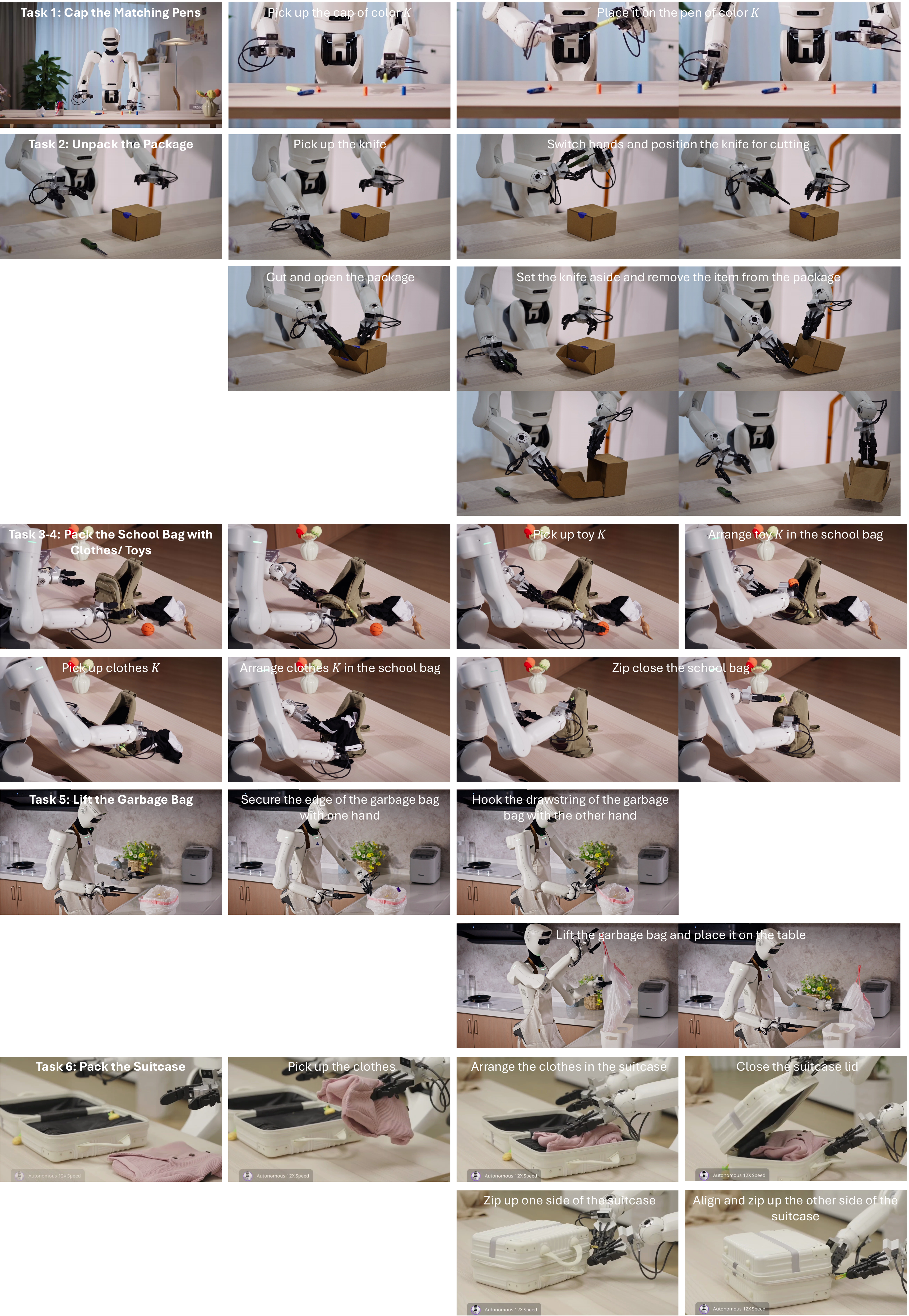}
   \caption{\textbf{Representative policy rollouts for Tasks 1-6.} For each task, the keyframes are displayed in the order of the subtasks as defined in Tab.~\ref{tab:task_list}.
}
\label{fig:task_illustration_1_6}
\end{figure}

\begin{figure}[p]
  \centering
  \includegraphics[width=0.95\linewidth]{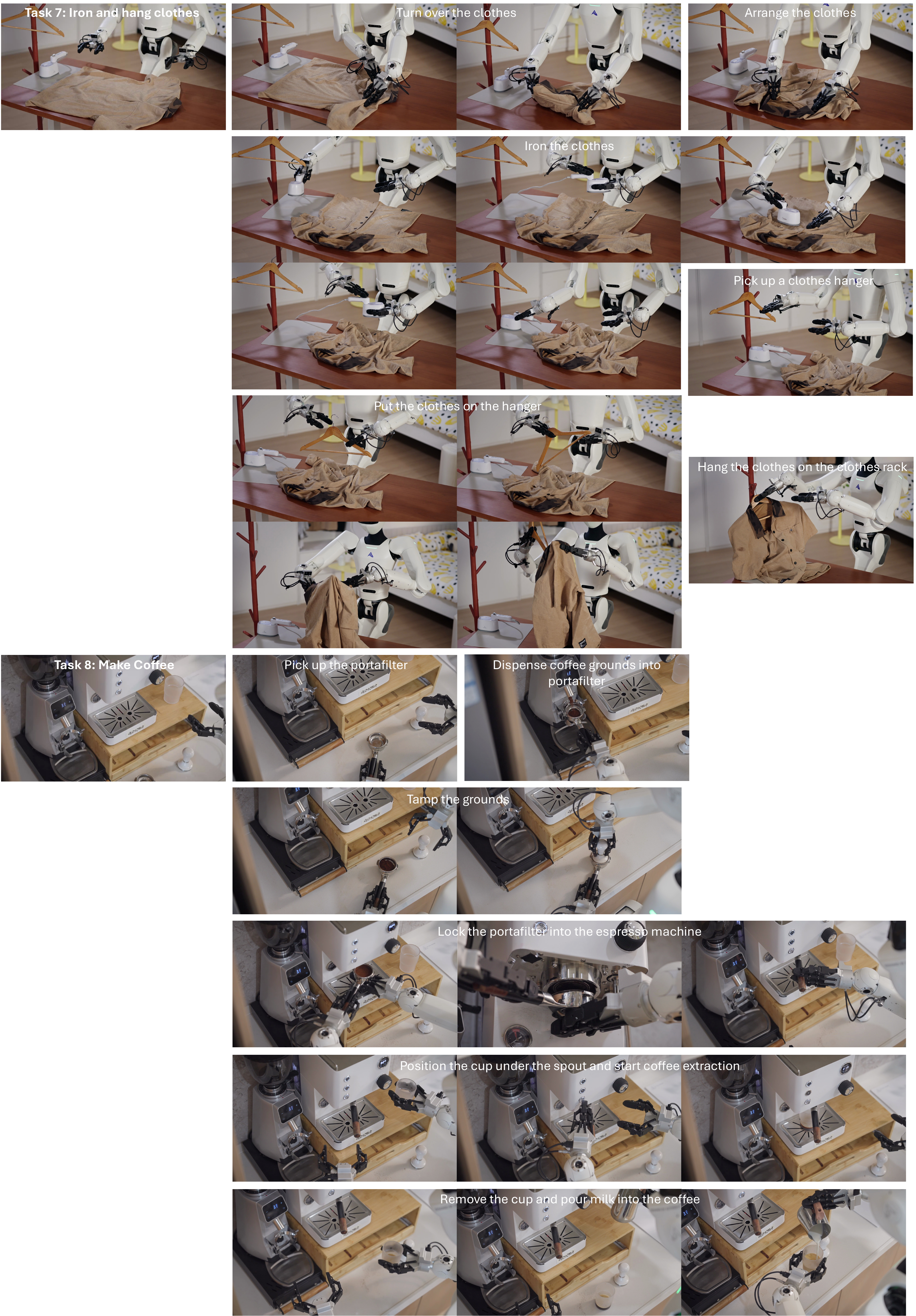}
  \caption{\textbf{Rollout illustrations for Tasks 7-8.} For each task, keyframes are depicted in the temporal order of the subtasks as enumerated in Tab.~\ref{tab:task_list}.}
  \label{fig:task_illustration_7_8}
\end{figure}

\begin{figure}[p]
  \centering
  \includegraphics[width=\linewidth]{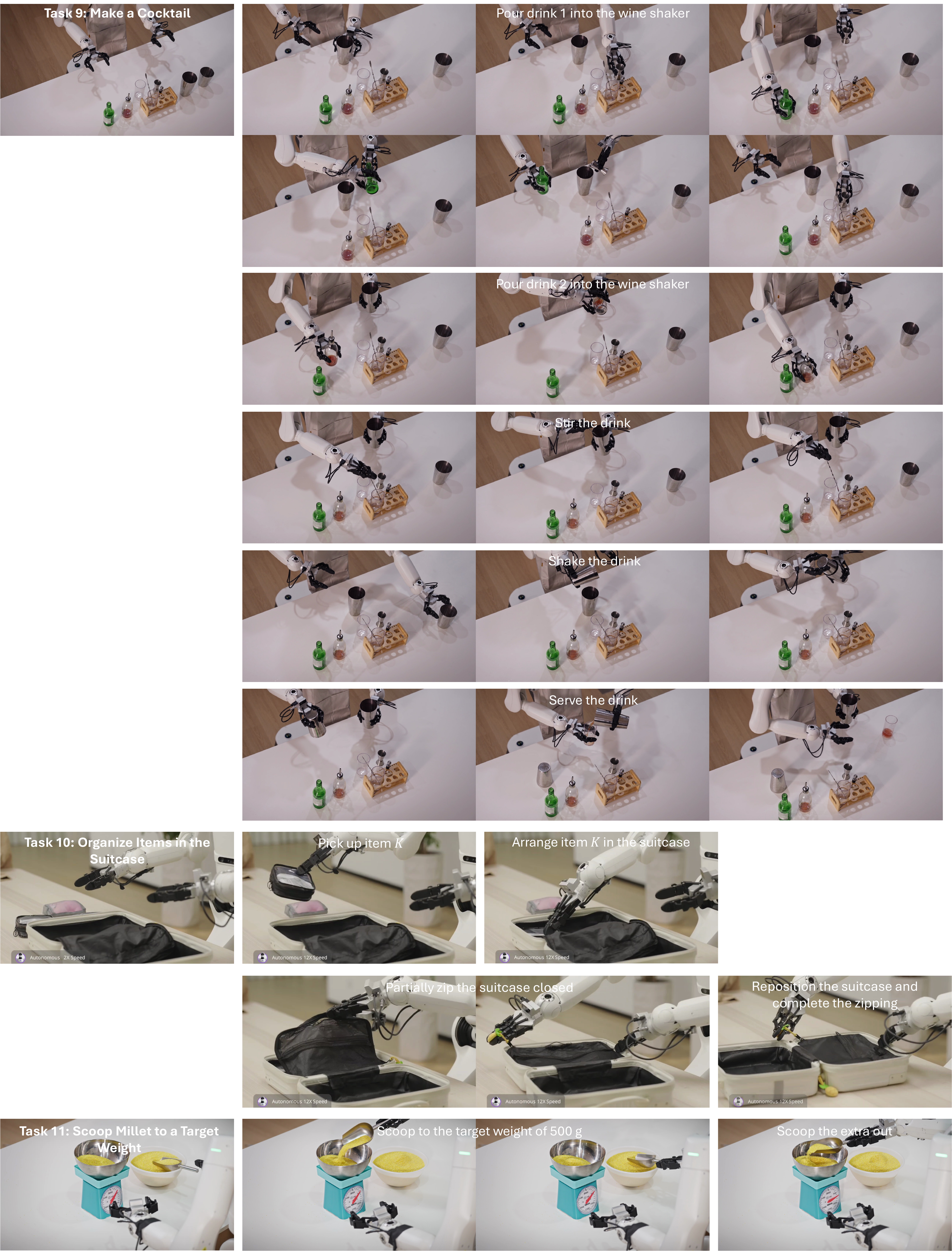}
  \caption{\textbf{Visualized policy rollouts for Tasks 9-11.} Within each task, the keyframes are shown in the sequential order of the subtasks specified in Tab.~\ref{tab:task_list}.}
  \label{fig:task_illustration_9_11}
\end{figure}

\begin{figure}[p]
  \centering
  \includegraphics[width=\linewidth]{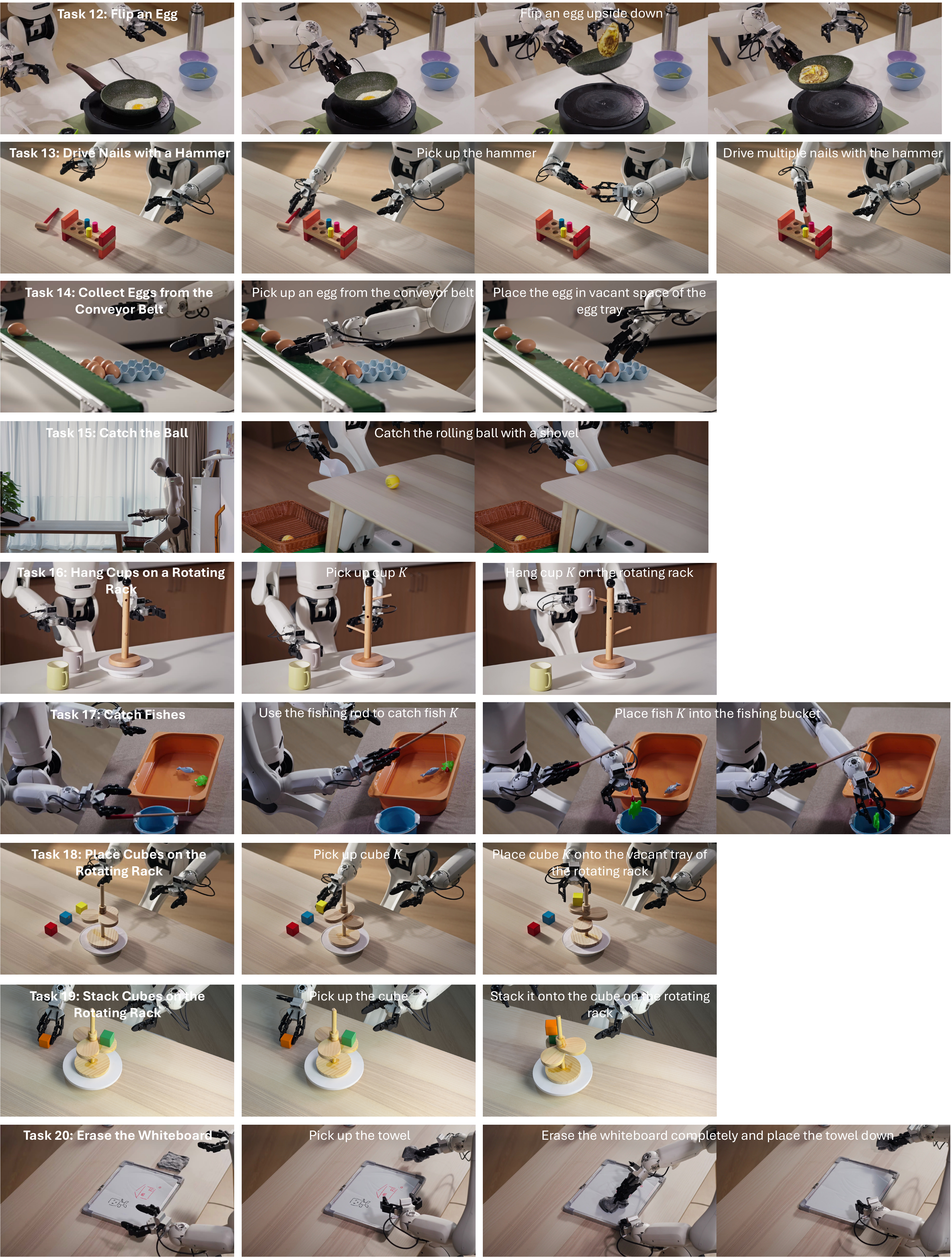}
  \caption{\textbf{Per-task rollout keyframes for Tasks 12-20.} The frame order for each task corresponds to the subtask breakdown provided in Tab.~\ref{tab:task_list}.}
  \label{fig:task_illustration_12_20}
\end{figure}

\begin{figure}[t!]
  \centering
  \includegraphics[width=\linewidth]{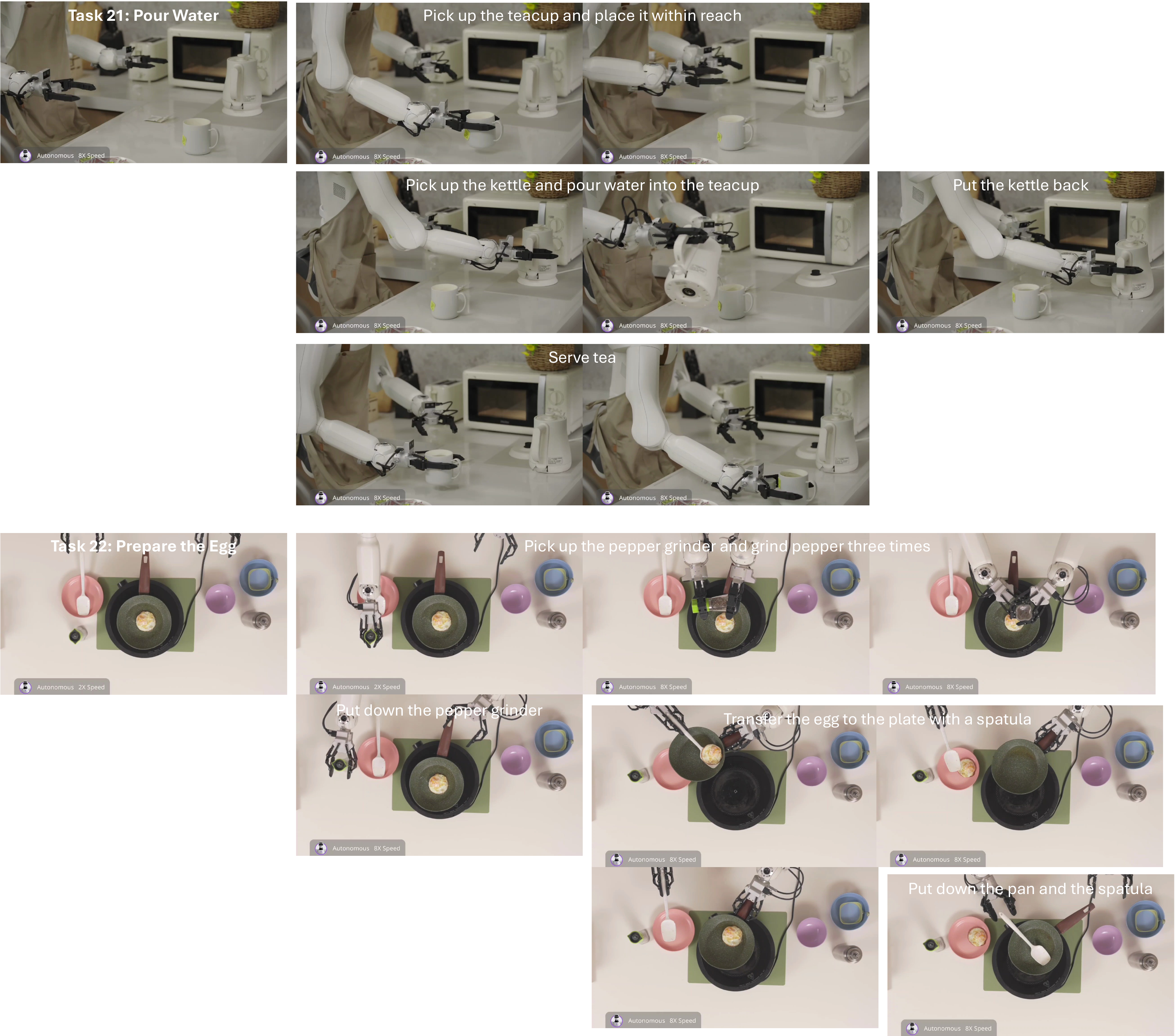}
  \caption{\textbf{Rollout sequences for Tasks 21-22.} For every task, the displayed frames follow the exact subtask order defined in Tab.~\ref{tab:task_list}.}
  \label{fig:task_illustration_21_22}
\end{figure}

We evaluate on a diverse suite of challenging manipulation tasks, as summarized in Tab.~\ref{tab:task_list}. Critical phases of each task are highlighted in red. Figs.~\ref{fig:task_illustration_1_6} -- \ref{fig:task_illustration_21_22} presents representative policy rollouts, highlighting the key stages of each task.

\subsubsection{Baselines and Evaluation Protocol.}

\begin{figure}[htb!]
  \centering
  \includegraphics[width=\linewidth]{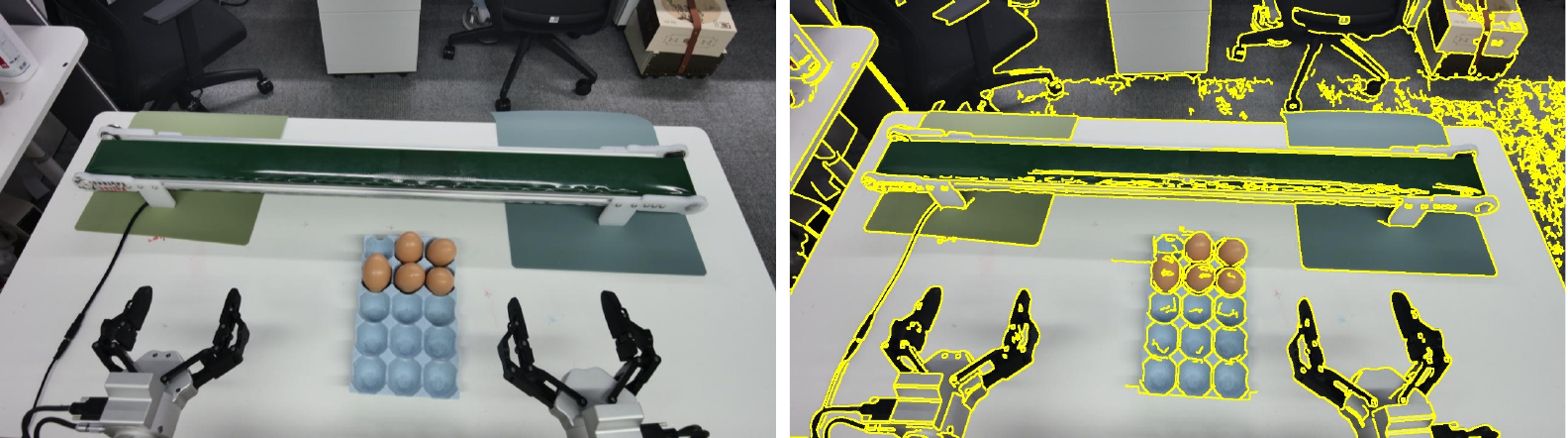}
    \caption{\textbf{Ensuring Consistent Evaluation Setups.} To enable fair comparisons across models and trials, we record the scene layout and overlay edge maps extracted from the robot's head-camera observations with a reference image, ensuring consistent object placement and scene configuration during evaluation.}
\label{fig:evaluation_protocol}
\end{figure}

We compare \model against two state-of-the-art baselines, $\pi_{0.5}$~\citep{pi_0_5} and Fast-WAM~\citep{yuan2026fast}, representing vision-language-action (VLA) and world-action model (WAM) approaches, respectively. For each task, we predefine 10 scene layouts with distinct initial conditions. To ensure a fair comparison, evaluation conditions are carefully controlled such that all models receive identical head-camera observations of the relevant objects, as illustrated in Fig.~\ref{fig:evaluation_protocol}. Each layout is evaluated twice to account for inference stochasticity, resulting in 20 blind evaluation trials per model. Importantly, the evaluation set intentionally includes diverse and challenging scene configurations to assess the models' generalization capabilities beyond the training distribution. We observe that Fast-WAM exhibits high sensitivity to visual input quality. To eliminate visual distribution gaps and guarantee equitable comparison, we apply dedicated image preprocessing for this baseline, including image augmentation, resolution upscaling, and aspect ratio adjustment matched to our robot camera observations. Note that many tasks are inherently long-horizon, even when categorized under a different primary challenge. For evaluation, we group multiple low-level subtasks into semantically coherent meta-subtasks, without distinguishing individual primitive actions such as ``pick up'' and ``put down''.

The evaluation metric is a normalized score. For single-step tasks, performance is measured on a binary scale (0–1), with 1 indicating success. For multi-step tasks, one point is awarded for completing each subtask, and the final score is averaged over the number of subtasks. Partial completion yields fractional scores to reflect progress. Category-level averages are computed across all tasks sharing the same primary challenge as specified in Tab.~\ref{tab:task_categories}. Some tasks contain particularly challenging subtasks. For these tasks, we evaluate performance at the subtask level. Each subtask starts from an in-domain initial configuration, and success is defined by the correct execution of the commanded subtask. We mark these tasks with an asterisk.

\begin{table}[]
\renewcommand{\arraystretch}{1.25} 
\resizebox{0.97\linewidth}{!}{%
\begin{tabular}{|l|c|c|l|}
\hline
Task ID & Task                      & Primary Challenge       & \multicolumn{1}{c|}{Subtask Sequence}                \\ \hline

1       & Cap the Matching Pens & Dexterous Manipulation &
\begin{tabular}[c]{@{}l@{}}
$
\left[
\begin{array}{l}
\text{Pick up the cap of color $K$.}\\
\rowcolor{red!15} \text{Place it on the pen of color $K$.}
\end{array}
\right]
\times 3
$ 
\end{tabular}
\\ \hline

2       & Unpack the Package            & Dexterous Manipulation &  \multicolumn{1}{l|}{\begin{tabular}[c]{@{}l@{}}
1. Pick up the knife.\\ 
2. Switch hands and position the knife for cutting. \\ 
\rowcolor{red!15} 3. Cut and open the package.\\ 
\rowcolor{white} 4-5. Set the knife aside and remove the item from the package. \end{tabular}} 
\\ \hline

3       & Pack the School Bag with Clothes*          & Dexterous Manipulation & \multicolumn{1}{l|}{\begin{tabular}[c]{@{}l@{}} 
$\left[
\begin{array}{l}
\text{Pick up clothes $K$.}\\
\text{Arrange clothes $K$ in the school bag.}
\end{array}
\right] \times 2$ 
\\ \rowcolor{red!15} 5. Zip close the school bag.  \end{tabular}} 
\\ \hline

4       & Pack the School Bag with Toys            & Dexterous Manipulation & \multicolumn{1}{l|}{\begin{tabular}[c]{@{}l@{}} 
$\left[
\begin{array}{l}
\text{Pick up toy $K$.}\\
\text{Arrange toy $K$ in the school bag.}
\end{array}
\right] \times 3$ 
\\ \rowcolor{red!15} 7. Zip close the school bag.  \end{tabular}} 
\\ \hline

5   & Lift the Garbage Bag            & Dexterous Manipulation & 
\multicolumn{1}{l|}{\begin{tabular}[c]{@{}l@{}}
1. Secure the edge of the garbage bag with one hand.\\ 
\rowcolor{red!15} 2. Hook the drawstring of the garbage bag with the other hand.\\ 
\rowcolor{white} 3. Lift the garbage bag and place it on the table.\\
\end{tabular}} 
\\ \hline

6   & Pack the Suitcase*            & Dexterous Manipulation & 
\multicolumn{1}{l|}{\begin{tabular}[c]{@{}l@{}}1-2. Pick up the clothes and arrange the clothes in the suitcase.\\ 3. Close the suitcase lid. \\ 
\rowcolor{red!15} 4. Zip up one side of the suitcase. \\ 5. Align and zip up the other side of the suitcase.  \end{tabular}} 
\\ \hline

7 &  Iron and Hang Clothes* & Dexterous Manipulation  & 
\multicolumn{1}{l|}{\begin{tabular}[c]{@{}l@{}}
\rowcolor{red!15} 1-2. Turn over and arrange the clothes.\\
\rowcolor{red!15} 3. Iron the clothes.\\
\rowcolor{red!15} 4-5. Pick up a clothes hanger and put the clothes on the hanger.\\
\rowcolor{white} 6. Hang the clothes on the clothes rack.\\
\end{tabular}} 
\\ \hline

8   & Make Coffee*        & Long-Horizon & 
\multicolumn{1}{l|}{\begin{tabular}[c]{@{}l@{}}
\rowcolor{red!15} 1-2. Pick up the portafilter and dispense coffee grounds into it.\\ 
\rowcolor{red!15} 3. Tamp the grounds.\\
\rowcolor{red!15} 4. Lock the portafilter into the espresso machine.\\
\rowcolor{white} 5-6. Position the cup under the spout and start coffee extraction.\\
\rowcolor{white} 7. Remove the cup and pour milk into the coffee.\\
\end{tabular}} 
\\ \hline

9 & Make a Cocktail            & Long-Horizon & 
\multicolumn{1}{l|}{\begin{tabular}[c]{@{}l@{}}
1-2. Pour drink 1-2 into the wine shaker.\\ 
\rowcolor{red!15} 3-5. Stir the drink, shake the drink, and serve. \\
\end{tabular}} 
\\ \hline

10   & Organize Items in the Suitcase            & Physical Understanding &
 \multicolumn{1}{l|}{\begin{tabular}[c]{@{}l@{}} 
$\left[
\begin{array}{l}
\text{Pick up item $K$.}\\
\text{Arrange item $K$ in the suitcase.}
\end{array}
\right] \times 4$ 
\\ 9. Partially zip the suitcase closed.  
\\ 10. Reposition the suitcase and complete the zipping. \end{tabular}}
\\ \hline

11   & Scoop Millet to a Target Weight          & Physical Understanding & Scoop millet to a target weight.
\\ \hline

12   & Flip an Egg         & Physical Understanding & Flip an egg upside down.
\\ \hline

13   & Drive Nails with a Hammer          & Physical Understanding & 
\multicolumn{1}{l|}{\begin{tabular}[c]{@{}l@{}}
1. Pick up the hammer. \\
2. Drive multiple nails with the hammer.
\end{tabular}} 
\\ \hline

14 & Collect Eggs from the Conveyor Belt         & Dynamic Scene  & 
{\begin{tabular}[c]{@{}l@{}} 
$\left[
\begin{array}{l}
\rowcolor{red!15} \text{Pick up an egg from the conveyor belt.}\\
\rowcolor{white} \text{Place it in vacant space of the egg tray.}
\end{array}
\right] \times 5$ \\
\end{tabular}} 
\\ \hline

15 & Catch the Ball          & Dynamic Scene  &   Catch the rolling ball with a shovel.
\\ \hline

16  & Hang Cups on a Rotating Rack  & Motion Reasoning & 
{\begin{tabular}[c]{@{}l@{}} 
$\left[
\begin{array}{l}
\text{Pick up cup $K$.}\\
\rowcolor{red!15} \text{Hang cup $K$ on the rotating rack.}
\end{array}
\right] \times 2$ \\
 \end{tabular}} 
\\ \hline

17 & Catch Fishes         & Motion Reasoning & 
{\begin{tabular}[c]{@{}l@{}} 
$\left[
\begin{array}{l}
\rowcolor{red!15} \text{Use the fishing rod to catch fish $K$.}\\
\rowcolor{white} \text{Place fish $K$ into the fishing bucket.}
\end{array}
\right] \times 3$ \\
 \end{tabular}} 
\\ \hline

18  & Place Cubes on the Rotating Rack        & Motion Reasoning & 
{\begin{tabular}[c]{@{}l@{}} 
$\left[
\begin{array}{l}
\text{Pick up cube $K$.}\\
\rowcolor{red!15} \text{Place cube $K$ onto the vacant tray of the rotating rack.}
\end{array}
\right] \times 3$ \\
 \end{tabular}} 
\\ \hline

19  & Stack Cubes on the Rotating Rack        & Motion Reasoning &  Pick up the cube and stack it onto the cube on the rotating rack.     
\\ \hline

20       & Erase the Whiteboard              & Memory &  \multicolumn{1}{l|}{\begin{tabular}[c]{@{}l@{}}1. Pick up the towel.\\ \rowcolor{red!15} 2. Erase the whiteboard completely and place the towel down.\end{tabular}} 
\\ \hline

21   & Pour Water           & Memory & 
\multicolumn{1}{l|}{\begin{tabular}[c]{@{}l@{}}
1. Pick up the teacup and place it within reach.\\ 
\rowcolor{red!15} 2. Pick up the kettle and pour water into the teacup.\\ 
\rowcolor{white} 3. Put the kettle back and serve tea.\end{tabular}} 
\\ \hline

22   & Prepare the Egg           & Memory & 
\multicolumn{1}{l|}{\begin{tabular}[c]{@{}l@{}}
\rowcolor{red!15} 1-2. Pick up the pepper grinder and grind pepper three times.\\ 
\rowcolor{white} 3. Put down the pepper grinder.\\
\rowcolor{red!15} 4. Transfer the egg to the plate with a spatula.\\
\rowcolor{white} 5. Put down the pan and the spatula.\\
\end{tabular}} 
\\ \hline
\end{tabular}
}
\caption{\textbf{Real-World Task Suites}. Each task is assigned one primary capability challenge. One point is awarded for completing each meta-subtask, so the overall task score naturally reflects the degree of completion.}
\label{tab:task_list}
\end{table}

\subsubsection{Results.}
\begin{figure}[t!]
  \centering
  \includegraphics[width=\linewidth]{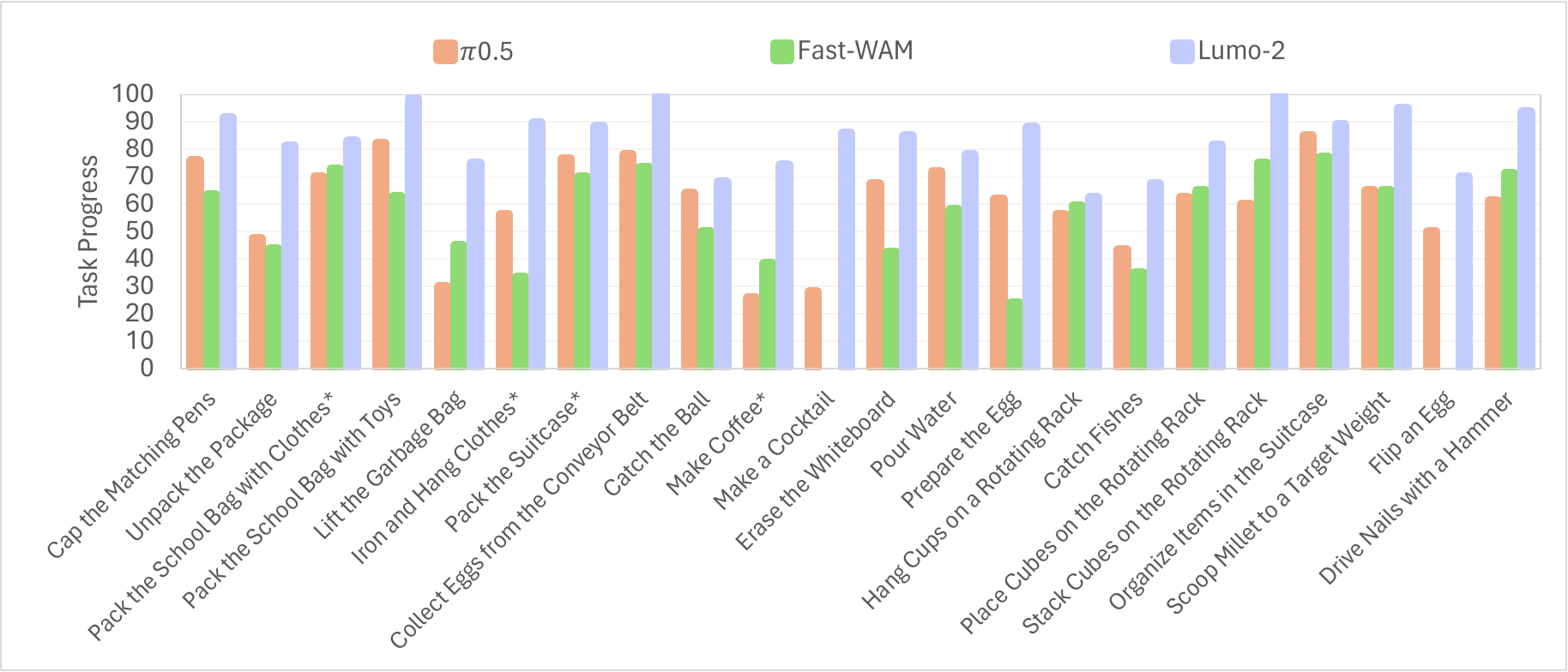}
  \caption{\textbf{Success rates on all 22 real‑world manipulation tasks.} Lumo‑2 consistently surpasses $\pi_{0.5}$ and Fast‑WAM, demonstrating comprehensive gains across the full evaluation suite.}
  \label{fig:task_results}
\end{figure}

\begin{figure}[t!]
  \centering
  \includegraphics[width=0.8\linewidth]{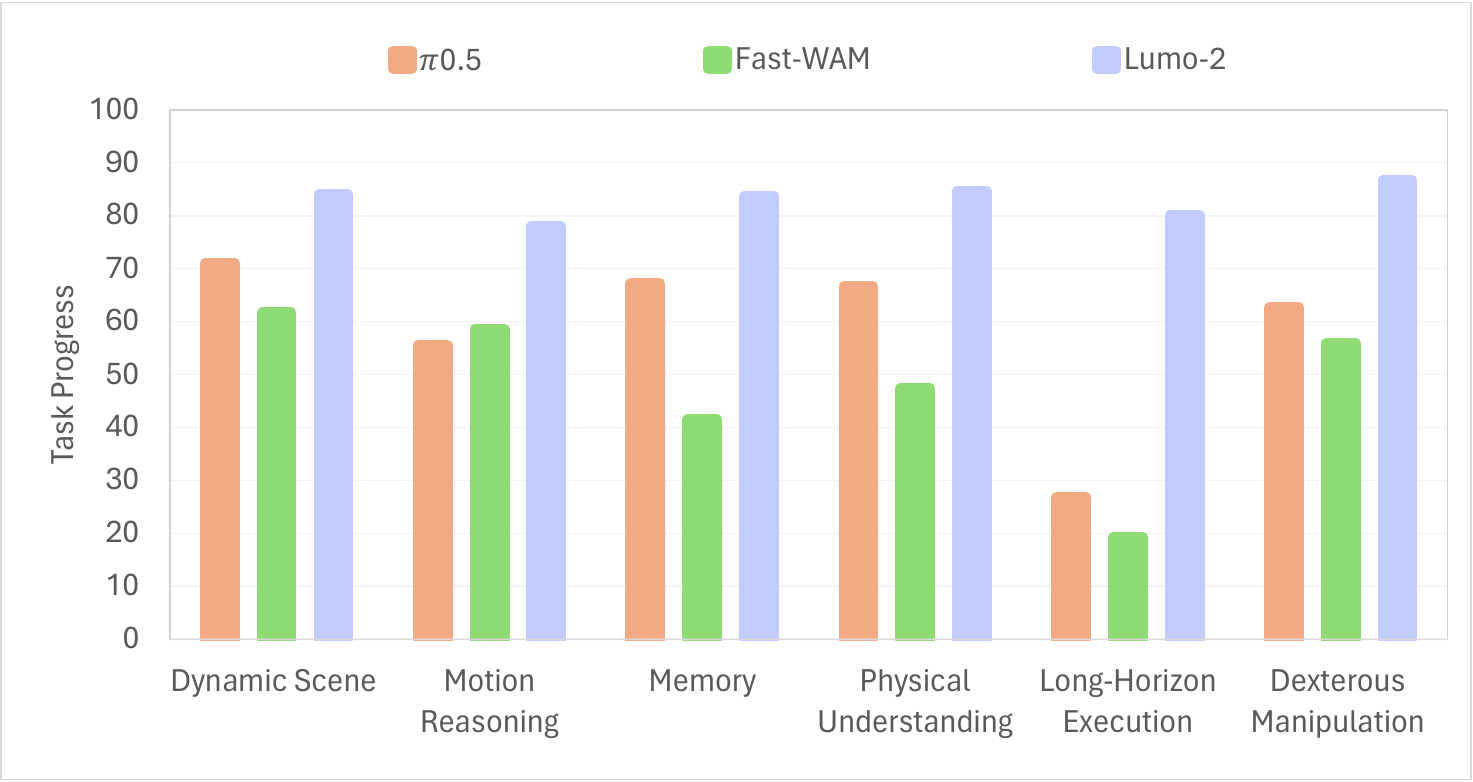}
  \caption{\textbf{Category‑level success rates.} Tasks are grouped into six sub‑categories: Dynamic Scene, Motion Reasoning, Memory, Physical Understanding, Long‑Horizon Execution, and Dexterous Manipulation. Lumo‑2 achieves superior performance in every category.}
  \label{fig:task_results_by_category}
\end{figure}

Detailed per-task success rates are reported in Fig.~\ref{fig:task_results}, and category‑level comparisons are provided in Fig.~\ref{fig:task_results_by_category}.
\model outperforms both baseline models $\pi_{0.5}$~\citep{pi_0_5} and Fast-WAM~\citep{yuan2026fast} across all task categories.

\paragraph{Strength 1: Temporal reasoning across reactive, predictive, and memory‑demanding tasks.}
\model consistently outperforms both baselines across all three sub‑categories of temporal reasoning.
For \textbf{dynamic scene reaction}, it handles \textit{Collect Eggs from the Conveyor Belt} and \textit{Catch the Ball} with markedly higher reliability, acting before transient interaction windows close.
In \textbf{motion reasoning}, \textit{Hang Cups on a Rotating Rack}, \textit{Catch Fishes}, \textit{Place Cubes on the Rotating Rack}, and \textit{Stack Cubes on the Rotating Rack} all show clear gains, with the latter two reaching near‑perfect execution, confirming the model's ability to anticipate spatiotemporal evolution.
\textbf{Memory}‑intensive tasks—\textit{Erase the Whiteboard}, \textit{Pour Water}, and \textit{Prepare the Egg}—also improve substantially over both baselines, demonstrating effective disambiguation of visually similar states across different execution phases.
These results validate that the latent world‑model and temporal context mechanism reliably resolve perceptual aliasing and preserve phase awareness.

\paragraph{Strength 2: Physical and causal understanding of interaction‑rich scenarios.}
\model demonstrates strong superiority on all tasks requiring reasoning about physical laws such as gravity, contact, and material behaviors.
On \textit{Organize Items in the Suitcase}, \textit{Scoop Millet to a Target Weight}, \textit{Flip an Egg}, and \textit{Drive Nails with a Hammer}, it achieves consistent and substantial improvements over the baselines.
Notably, Fast‑WAM~\citep{yuan2026fast} completely fails on \textit{Flip an Egg}. We find that its action space is too restricted to generate the large‑amplitude flipping motion required, likely because the action head lacks sufficient pre‑training on diverse physical interactions.
The consistent improvements confirm that \model can anticipate causal, often non‑linear interaction outcomes and adapt its motion accordingly, without explicit physics simulators.

\paragraph{Strength 3: Robust long‑horizon execution and high‑precision dexterous control.}
\model excels in both sub‑categories of control complexity.
For \textbf{long‑horizon execution}, \textit{Make Coffee} and \textit{Make a Cocktail} witness dramatic gains; Fast‑WAM~\citep{yuan2026fast} scores zero on the latter. Analysis shows that Fast‑WAM struggles to disambiguate visually similar intermediate states (e.g., whether the bottle has been picked up or the shaker has been emptied), causing it to skip or stall on critical steps and ultimately fail to chain the full sequence. \model's latent dynamics model, by explicitly predicting future states, maintains a clear phase memory and executes each step reliably.
In \textbf{dexterous manipulation}, \model achieves the best performance across all seven tasks—\textit{Cap the Matching Pens}, \textit{Unpack the Package}, \textit{Pack the School Bag with Clothes}, \textit{Pack the School Bag with Toys}, \textit{Lift the Garbage Bag}, \textit{Pack the Suitcase}, and \textit{Iron and Hang Clothes}—demonstrating reliable high‑precision control and tight‑tolerance bimanual coordination.
These results underscore \model's capacity to maintain coherence over extended horizons, recover from intermediate failures, and handle complex object geometries with contact stability.

\subsection{Human to Robot Transfer [$\mathcal{Q}4$]}

As robotic foundation models become more capable, we expect them to act as generalists, able to leverage diverse data sources in ways that smaller models cannot~\citep{kareer2025emergence}. Here, we investigate whether \model exhibits emergent capabilities for integrating new data sources. We focus specifically on human demonstration data collected from different devices and viewing setups, which can be acquired at scale with relative ease but differ substantially from standard robot data in view coverage and annotation completeness. Our goal is to determine whether \model can facilitate emergent human-to-robot transfer without any explicit transfer learning mechanism, and how the availability of world dynamics and action supervision affects downstream policy performance.

\begin{figure}[t!]
  \centering
  \includegraphics[width=\linewidth]{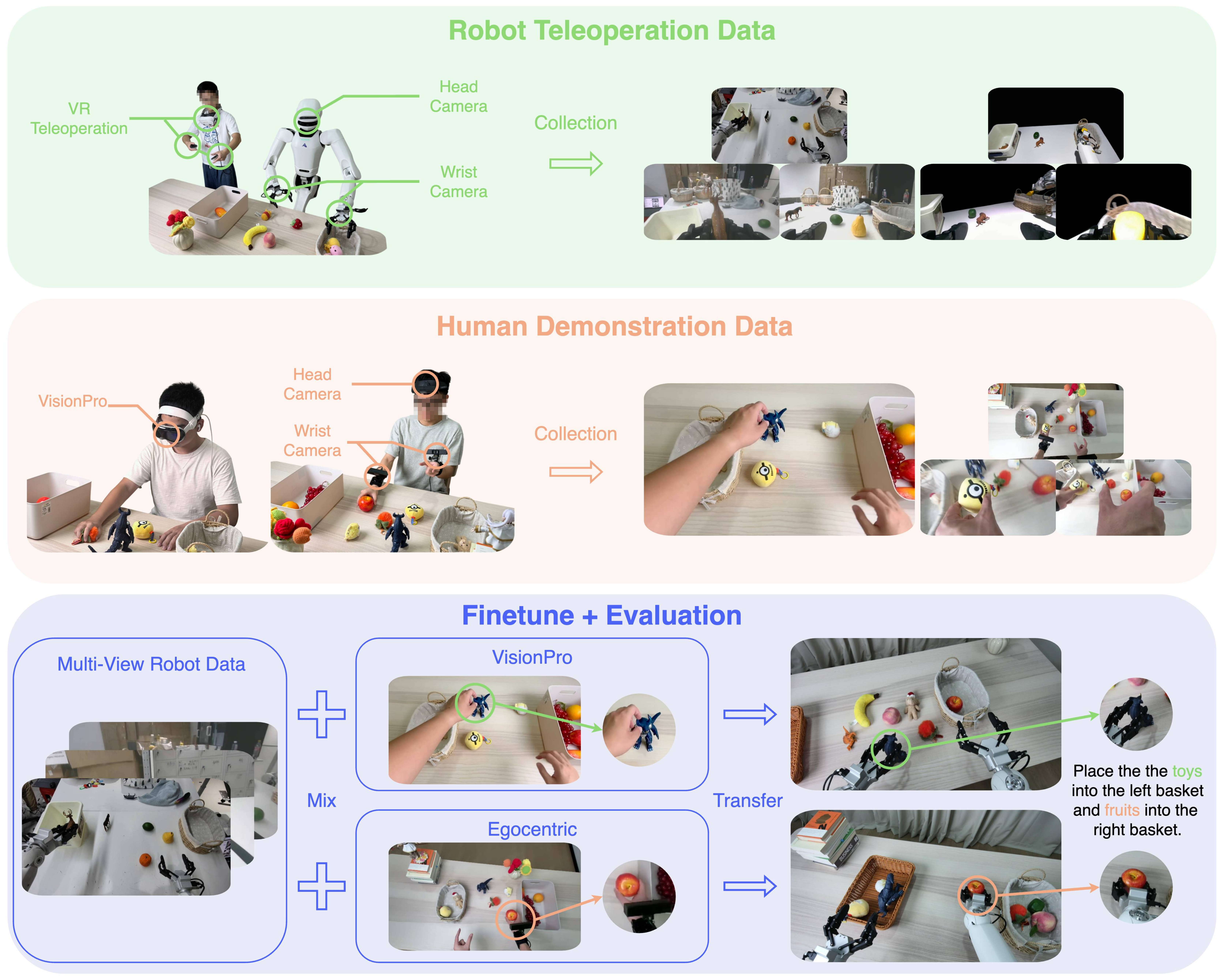}
  \caption{\textbf{Human-to-Robot Transfer Schematic.} Schematic overview of the emergent human-to-robot transfer framework, including data acquisition devices, data modalities, the human-robot co-finetuning pipeline, and evaluation on unseen robot objects, with no specialized transfer learning mechanisms employed.}
  \label{fig:human2robot}
\end{figure}

To this end, we employ a simple and unified human-robot co-finetuning protocol across all experiments, which requires no specialized transfer learning mechanisms. Our native robot dataset serves as the full-modality baseline, with complete three-view coverage and full-dimensional action annotations (including torso, bimanual poses, and gripper states) that support all training objectives. For each external human data source, we treat human demonstrations in the same way as our existing robot embodiments: we first perform targeted data cleaning and modality alignment to fit our model's unified input schema, then mix it with the most relevant robot data for joint fine-tuning, and finally evaluate the resulting policies on robot task settings that appear only in human demonstrations. We adopt different combinations of training objectives according to the view coverage and annotation completeness of each data source, covering both action-supervised and action-free modes. In the absence of action labels, training reduces to world dynamics understanding, which in turn exerts a positive effect on subsequent action learning. 

Fig.~\ref{fig:human2robot} provides a schematic overview of the view configurations, annotation modalities and corresponding training paradigms for all employed human demonstration datasets, alongside our native multi-view robot data used as the co-finetuning baseline. The following sections elaborate on the data characteristics, core processing pipelines, and evaluation setup for each data source.

\paragraph{VisionPro Demonstrations.}
We collect human manipulation demonstrations using an Apple VisionPro headset, which synchronously captures first-person RGB observations and 3D bimanual hand poses via the built-in ARKit tracking system.
We timestamp each data frame to ensure precise temporal alignment between visual and motion streams. For core processing, we perform kinematic retargeting to convert human hand poses into robot-compatible action labels: we first transform poses from the head-mounted coordinate system to the robot base frame, then map human wrist poses to the robot’s dual-arm end-effectors, and infer gripper commands from the distance between thumb and index finger. The torso dimension is padded with fixed values to match our full action space definition.
Despite the single viewpoint, the first-person observations provide sufficient spatial cues for world dynamics modeling. We therefore train this data under the full vision-language-world dynamics-action (VLWA) objective and co-finetune it with native robot data.

\paragraph{Multi-View Egocentric Videos.}
This dataset provides three synchronized views of human manipulation with sufficient spatial coverage for world dynamics inference, but contains no paired action labels. We therefore train under the vision-language-world dynamics (VLW) objective only, where the model learns scene evolution and physical states from multi-view observations without action supervision. We co-finetune this data with our full robot dataset to verify whether learned world dynamics alone can improve downstream robot control.

\begin{figure}[t!]
  \centering
  \includegraphics[width=0.8\linewidth]{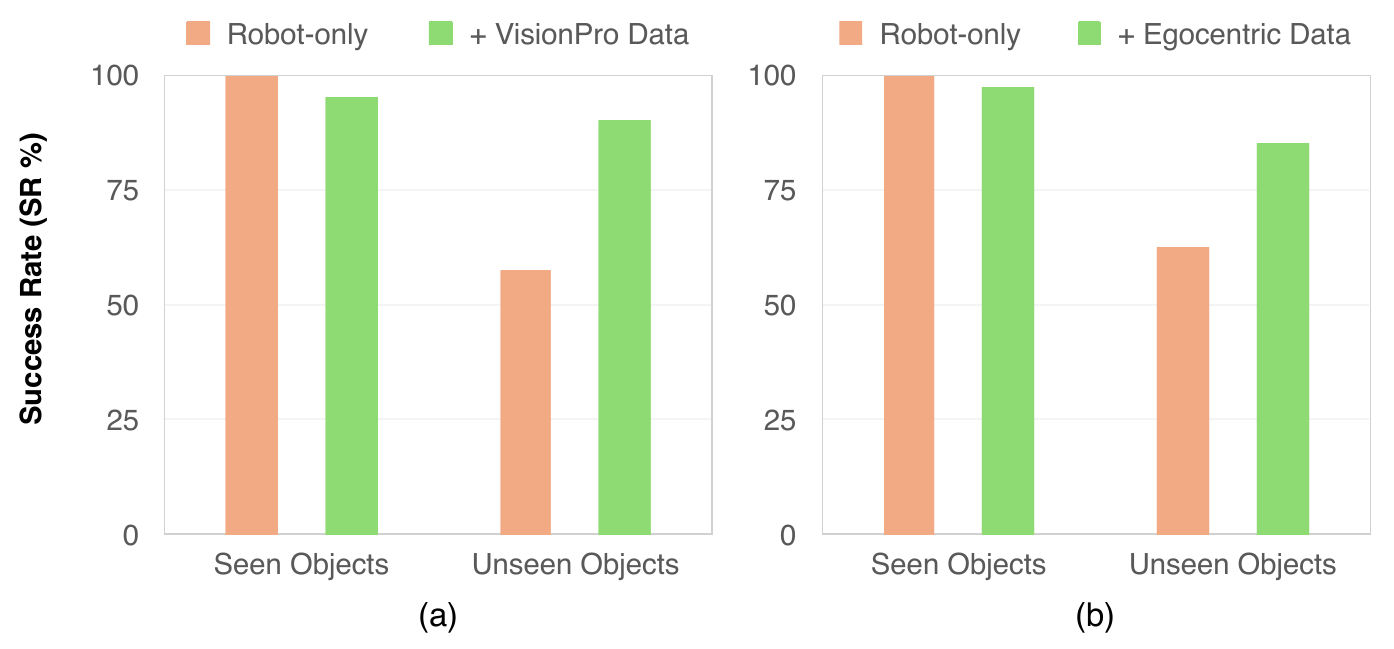}
  \caption{\textbf{Human-to-Robot Transfer Evaluation Results.} As shown in (a) and (b), we compare pick-and-place success rates between the robot-only baseline and the co-trained model on seen and unseen objects (defined relative to the native robot training set). Both human data sources deliver significant improvements on unseen objects without requiring specialized transfer learning mechanisms.}
  \label{fig:human2robot_evaluate}
\end{figure}

\paragraph{Evaluation Protocol.}
To ensure fair comparison, all experiments use strictly matched training steps and hyperparameters. For each human data source, we compare two model variants: a baseline fine-tuned solely on native robot data, and a co-trained model fine-tuned on mixed robot and human demonstration data.
We evaluate on a standard pick-and-place task, with objects split into two categories relative to the native robot training set: \textit{seen} and \textit{unseen}. A trial is counted as successful if the target object is successfully picked up and placed at the designated position. We run repeated trials with randomized object positions for robust statistics, and report the average success rate for each category.
As summarized in Fig.~\ref{fig:human2robot_evaluate}, adding human demonstration data yields substantial performance gains on unseen objects for both data sources while maintaining competitive performance on seen objects, verifying the emergent cross-domain transfer capability of \model.

\subsection{Alignment and Scaling Laws [$\mathcal{Q}5$]}

\paragraph{Evaluation of Action Reconstruction Error.}

To validate the effectiveness of the proposed action tokenizer, we treat action reconstruction fidelity as the primary evaluation metric. For fine-grained robotic manipulation tasks, the encoding and decoding error of action sequences directly governs the model’s ability to learn actionable control signals and execute tasks successfully in the real world. To establish a fair baseline, we train an Action Only model for 30k steps with identical data volume and training configurations, which optimizes solely the discrete action autoencoding objective without any auxiliary supervision from vision, semantics, or world dynamics. We then compare its reconstruction performance with our Stage~2 checkpoint on an unseen test split of approximately 20k samples drawn from \model SFT data. All errors are computed in the decoded physical space to enable direct comparison of real-world control accuracy.

As summarized in Tab.~\ref{tab:action_reconstruction}, Stage~2 achieves lower reconstruction errors than the Action Only baseline across all SO(3) rotation and Cartesian position dimensions, with the only exception being the gripper channels. For instance, the torso geodesic error is nearly halved from 0.66° to 0.34°, and the left and right arm geodesic errors decrease by around 14\% and 17\% respectively. The positional MAE of both arm endpoints also shows consistent improvement. This result demonstrates that introducing world dynamics tokens and semantic information does not compromise action reconstruction quality. On the contrary, the additional contextual and structural signals help the tokenizer better model physical constraints and temporal dependencies embedded in action trajectories, leading to enhanced overall reconstruction accuracy. The slightly elevated gripper error in Stage~2 is acceptable in practice, as the gripper dimension intrinsically has a larger permissible error range and imposes less impact on most manipulation tasks.

\begin{table}[h]
\centering
\setlength{\extrarowheight}{3pt}
\resizebox{\linewidth}{!}{%
\begin{tabular}{ccccccccc}
\toprule
& Torso SO3
& Left SO3
& Right SO3
& Torso XYZ
& Left XYZ
& Right XYZ
& Left Gripper
& Right Gripper \\
& Geodesic (°) & Geodesic (°) & Geodesic (°)
& MAE (m) & MAE (m) & MAE (m) & MAE & MAE \\
\midrule
Action Only
& 0.66152 & 1.41548 & 1.59308
& \textbf{0.00038} & 0.00308 & 0.00367
& \textbf{0.40307} & \textbf{0.43429} \\
Stage~2
& \textbf{0.34059} & \textbf{1.22296} & \textbf{1.32877}
& 0.00043 & \textbf{0.00288} & \textbf{0.00359}
& 0.58081 & 0.59257 \\
\bottomrule
\end{tabular}
}
\caption{\textbf{Action reconstruction error comparison between the Action Only baseline and the Stage~2 checkpoint.} Results are evaluated on held-out Lumo-2 SFT data not seen during model training.}
\label{tab:action_reconstruction}
\end{table}

\begin{figure}[b!]
  \centering
  \includegraphics[width=\linewidth]{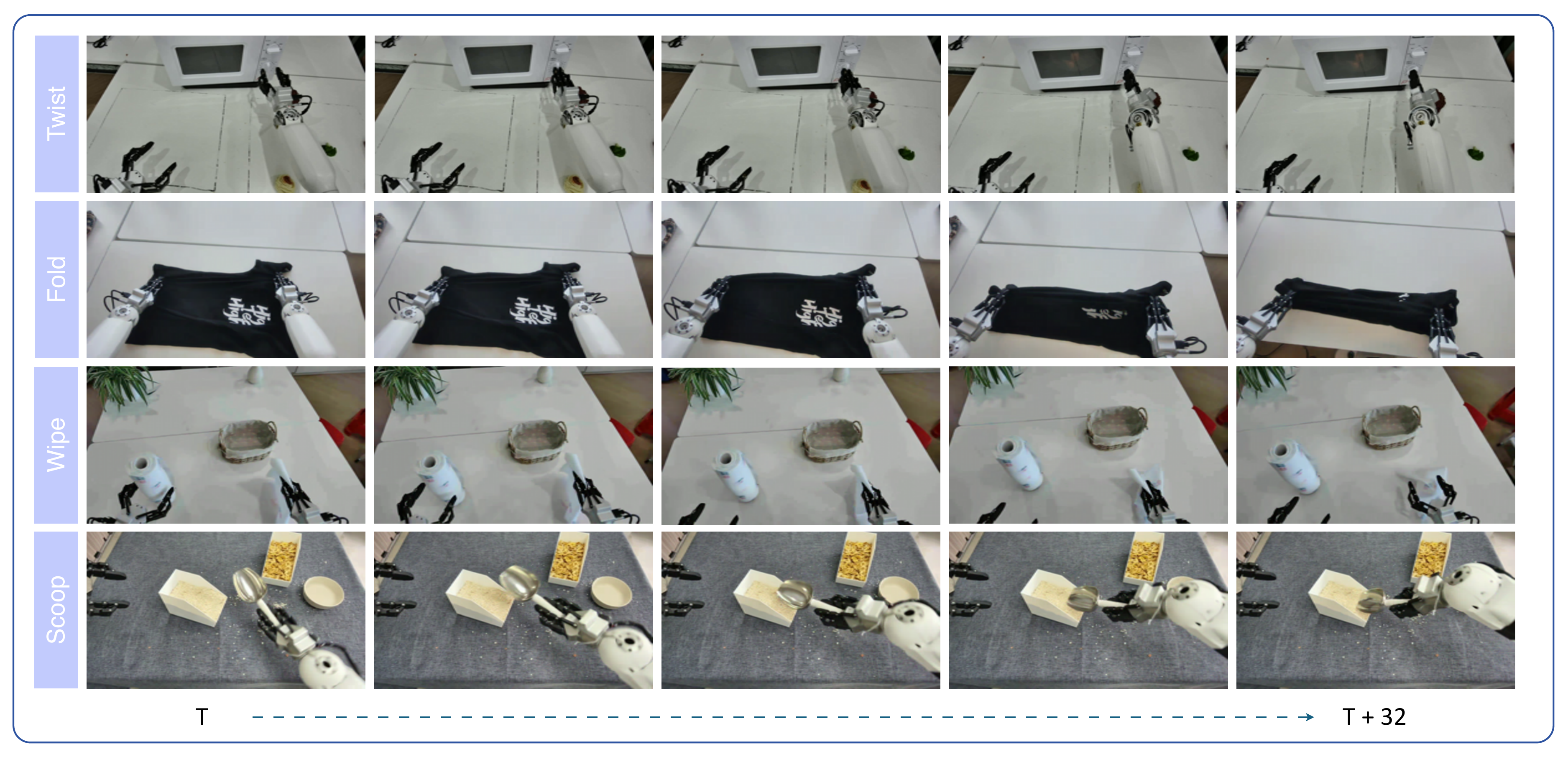}
   \caption{\textbf{Representative Examples of the Semantic Action Prediction Task.} Each row corresponds to an image sequence sampled at intervals of one action chunk (32 timesteps). The model is tasked with predicting the atomic action category based on the first frame and its corresponding action representation.}
   \label{fig:stage2_semantic_example}
\end{figure}

\paragraph{Evaluation of Action Semantic Prediction.}
We quantitatively evaluate the semantic discriminative capability of action representations from different training stages. Following the protocol of ~\citep{nie2026lary}, we construct an atomic action dataset based on our robot manipulation data. Specifically, we classify fine-grained task prompts using a large language model and apply heuristic data cleaning. From the processed data, we select 18 atomic action categories with sufficient sample volume and trajectory diversity, including \textbf{place, pick, grab, fold, open, pour, pull, twist, scoop, insert, cover, hang, wipe, peel, deliver, press, water}, and \textbf{mix}. Each category contains 400 samples, yielding a total of 7,200 samples. We split the dataset into training and test sets at a 4:1 ratio, with illustrative samples shown in Fig.~\ref{fig:stage2_semantic_example}. The task is to predict the semantic categories of atomic actions given the initial frame and the corresponding action representation. We conduct all evaluations under the linear probing paradigm. The classifier adopts a two-layer MLP architecture with a hidden dimension of 1024, ReLU activation, and a dropout rate of 0.1. 

As presented in the Tab.~\ref{tab:semantic_classify}, the DINO Only baseline extracts single-frame image features via a pre-trained DINOv2 encoder, serving as the performance reference for vision-only classification. We further build two comparative variants by augmenting the base visual features with Stage~1 action features and Stage~2 semantic action features, respectively. We use top-1 classification accuracy as the evaluation metric. The vision-only baseline achieves an accuracy of $84.10\%$, as visual cues from diverse objects already provide strong discriminative information. With the addition of Stage~1 action representations, the classification accuracy rises to $92.85\%$. When incorporating Stage~2 semantic action representations as supplementary input, the model attains the highest accuracy of $95.00\%$.

\begin{table}[htbp]
\centering
\begin{tabular}{ccc}
\toprule
DINO Only (Baseline) & Stage 1 Action + DINO & Stage 2 Semantic Action + DINO \\
\midrule
84.10 & 92.85 & \textbf{95.00} \\
\bottomrule
\end{tabular}
\caption{\textbf{Classification Results of Action Semantic Prediction with Different Representations.}}
\label{tab:semantic_classify}
\end{table}

\paragraph{Scaling with Semantic Action Representation.}
During Stage~2, we internalize latent world dynamics into the action decoder and pre-align the embedding spaces across action, vision, and language modalities. To verify the effectiveness of this pre-alignment for downstream action generation training, we conduct controlled experiments with identical training data and hyperparameters. The only variable between the two experimental groups is the action tokenizer adopted (from Stage~1 or Stage~2), along with minor differences in the action decoding workflow.

\begin{figure}[t!]
  \centering
  \includegraphics[width=0.98\linewidth]{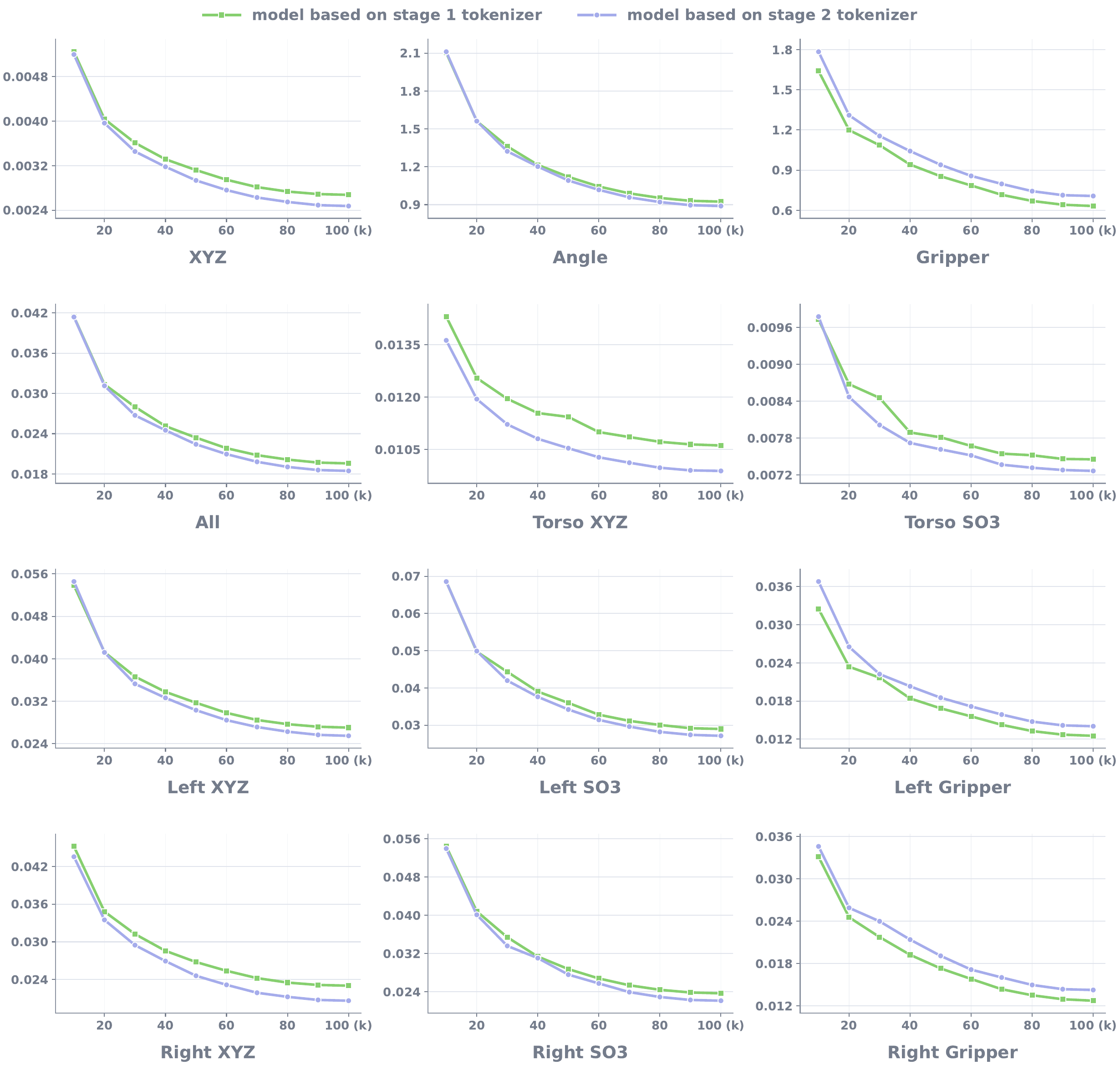}
  \caption{\textbf{Decoding error comparison across action tokenizers.}
  \textit{Top}: physical reconstruction errors vs. ground truth, including xyz position, angular error, and gripper closure error.
  \textit{Rows 2--4}: MAE of decoded outputs in normalized space, covering overall error and sub-errors for eight action groups.
  All subplots share the x-axis of training steps. The Stage~2 tokenizer consistently yields lower errors than Stage~1 on all metrics except gripper, which has inherently larger error tolerance.}
   \label{fig:only_pnp_absolute_and_delta}
\end{figure}

Both model variants are trained for 100,000 steps on 8 H100 GPUs with a per-device batch size of 40, using the General Pick and Place dataset from Lumo-1 under identical training configurations. We observe that for the BAR loss, the Stage~2-initialized model achieves lower loss in the early training phase, but converges to a higher final loss level than the Stage~1-based model as training proceeds. We hypothesize that this pattern arises because the Stage~2 VQ codebook encodes richer semantic information and supports finer-grained action discrimination. 
To further validate this hypothesis, we sample 100 complete trajectories (approximately 15,000 frames in total) as the validation set and perform frame-by-frame inference. We compare the two models in terms of mean absolute error (MAE) on two forms of decoded outputs: the normalized values directly produced by the action tokenizer, and the corresponding physical values converted via denormalization and transformation. As illustrated in Fig.~\ref{fig:only_pnp_absolute_and_delta}, the Stage~1-based model consistently shows larger overall reconstruction errors than the Stage~2-based model throughout the training process. This result empirically verifies the effectiveness of the lightweight pre-alignment scheme.

\section{Related Work}
\paragraph{Vision-Language-Action Models.}
Recent advances in robotic manipulation have marked a clear shift from narrowly specialized, single-task policies toward generalist models trained on large-scale, heterogeneous data. In this context, vision–language–action (VLA) models have shown notable improvements in both manipulation performance and policy generalization. Prior work broadly follows three training paradigms. (1) Incorporating pre-trained vision–language models (VLMs)~\citep{bai2025qwen2,beyer2024paligemma,qwen3vl} into control pipelines, enabling effective transfer to novel tasks and environments~\citep{brohan2022rt,rt_2,driess2023palm,pi_0,pi_0_5}. (2) Scaling data through large, cross-embodiment datasets that span diverse robot platforms and task distributions~\citep{o2024open,team2024octo,kim2024openvla,pi_0,pertsch2025fast,pi_0_5,liu2024rdt,li2024cogact}. (3) Joint multi-modal training to improve perception–action grounding and reasoning, often augmented with external vision foundation models that provide auxiliary action guidance~\citep{brohan2022rt,pi_0_5,li2025controlvla}.
To further enhance high-level reasoning, a growing body of work introduces intermediate representations or explicit reasoning processes within VLA frameworks. For example, ECoT~\citep{zawalski2024robotic} employs supervised fine-tuning to encourage reasoning prior to action execution, while CoT-VLA~\citep{zhao2025cot} replaces textual chain-of-thought with visually grounded subgoal generation. Other approaches, such as MolmoAct~\citep{lee2025molmoact} and Emma-X~\citep{sun2024emma}, autoregressively generate structured reasoning signals, including subtask decomposition, depth estimation, and future gripper states. ThinkAct~\citep{huang2025thinkact} further integrates action-aligned reinforcement learning with latent visual planning, aiming to better align embodied reasoning with action prediction. In our prior work, Lumo-1~\citep{lumo1}, we construct structured reasoning traces that jointly capture subtask decomposition, planning, and trajectory prediction. Despite these advances, action inference that conditions solely on the current observation - without explicitly modeling how the environment evolves under interaction - limits anticipatory reasoning and undermines long-horizon decision-making in dynamic, real-world settings.

\paragraph{World-Action Models.}
Recent studies have explored video generation as a surrogate for world modeling in robotic control, leveraging its ability to capture temporal dynamics and plausible future evolution, beyond the static perception and general understanding capability of vision-language models.  One line of work employs video models as predictive representation learners or transferable world priors, typically followed by separate action decoding modules~\citep{hu2024video,pai2025mimic,feng2025vidar,liao2025genie}. Another line of research pursues tighter integration by jointly modeling future visual observations and actions within a unified framework~\citep{li2025unified,zhu2025unified,liang2025video}, showing that co-predicting visual futures and action trajectories can improve both generalization and data efficiency.
Building on increasingly capable pretrained video foundation models~\citep{wan2025wan,agarwal2025cosmos}, more recent efforts move toward unified world–action formulations that support closed-loop control, causal rollout, and planning over predicted futures~\citep{bi2025motus,shen2025videovla}. Representative examples, including DreamZero~\citep{dreamzero} and LingBot-VA~\citep{lingbotva}, demonstrate that strong video priors can be effectively transferred into embodied policies, improving generalization and cross-embodiment transfer. Fast-WAM~\citep{yuan2026fast} further shows that maintaining video co-training during training -  while omitting explicit future generation at inference - preserves strong action performance with substantially reduced latency.
Our work is closely related to this line of research but departs in a fundamental way. Rather than relying on explicit future video rollout, we use future information to structure a latent reasoning space that directly guides action generation. While a complementary line of work explores future prediction or latent-space alignment for action learning~\citep{luo2025learning,vujinovic2025act,sun2026vla,zhang2026clap,chen2026unit,lyu2026lda,team2026being,Zhao2026FRAPPEIW,Su2026WorldGW,zheng2025flare,Maes2026LeWorldModelSE,Zhou2024DINOWMWM}, we instead explicitly and progressively pre-align actions with confounding modalities: a learned world dynamics space, vision, and language, yielding a more flexible and expressive reasoning substrate. This formulation promotes reasoning as latent chain-of-thought, where internal representations are jointly optimized to both predict future evolution and remain inferable from the current context. 

\section{Conclusion}
\label{sec:conclusion}
We introduce \model, a latent world–action model that unifies direct action prediction with world modeling through a compact latent reasoning space grounded in learned visual dynamics. This design injects future-aware reasoning into action generation without requiring expensive pixel-level rollouts at inference time. By progressively aligning the action modality with visual world dynamics, vision, and language, the model exhibits improved scaling behavior for action prediction. Coupled with large-scale pretraining on VLM, video, and robotic datasets, \model offers an effective and scalable framework for embodied intelligence.
\newpage
\section{Contributions}
Author contributions in the following areas are listed in alphabetical order.
\label{sec:contribution}
\begin{itemize}
    \item \textbf{Data:} Baifu Huang, Binyan Sun, Shangjin Xie, Shilin Fang
    \item  \textbf{Model Architecture:} Jianan Wang, Peijun Tang
    \item  \textbf{Training:} Baifu Huang, Binyan Sun, Haotian Yang, Kuncheng Luo, Peijun Tang, Shangjin Xie
    \item  \textbf{Evaluation:} Baifu Huang, Binyan Sun, Haotian Yang, Kuncheng Luo, Peijun Tang, Shangjin Xie,  Shilin Fang
    \item  \textbf{Writing:} Jianan Wang, Peijun Tang
    \item  \textbf{Project Lead:} Jianan Wang
\end{itemize}

\clearpage
\bibliography{main}

\end{document}